\documentclass[sn-mathphys,Numbered]{sn-jnl}

\usepackage{graphicx}%
\usepackage{multirow}%
\usepackage{amsmath,amssymb,amsfonts}%
\usepackage{amsthm}%
\usepackage{mathrsfs}%
\usepackage[title]{appendix}%
\usepackage{xcolor,xspace}%
\usepackage{textcomp}%
\usepackage{manyfoot}%
\usepackage{booktabs}%
\usepackage{algorithm}%
\usepackage{algorithmicx}%
\usepackage{algpseudocode}%
\usepackage{listings}%
\usepackage{subfig}%
\usepackage{enumitem}
\usepackage{xcolor}

\theoremstyle{thmstyleone}%
\theoremstyle{thmstyletwo}%

\theoremstyle{thmstylethree}%

\begin{document}

\title[Article Title]{Human-AI Collaboration in Real-World Complex Environment with Reinforcement Learning}


\author*[1,5]{\fnm{Md Saiful} \sur{Islam}}\email{mdsaifu1@ualberta.ca}
\author[1]{\fnm{Srijita} \sur{Das}}\email{srijita1@ualberta.ca}
\author[2]{\fnm{Sai Krishna} \sur{Gottipati}}\email{sai@ai-r.com}
\author[2]{\fnm{William} \sur{Duguay}}\email{william@ai-r.com}
\author[2]{\fnm{Clodéric} \sur{Mars}}\email{cloderic@ai-r.com}
\author[3]{\fnm{Jalal} \sur{Arabneydi}}\email{jalal.arabneydi@jacobb.ai}
\author[4]{\fnm{Antoine} \sur{Fagette}}\email{antoine.fagette@thalesgroup.com}
\author[1,5]{\fnm{Matthew } \sur{Guzdial}}\email{guzdial@ualberta.ca}
\author[1,5]{\fnm{Matthew E. } \sur{Taylor}}\email{matthew.e.taylor@ualberta.ca}

\affil*[1]{\orgname{University of Alberta}, \orgaddress{ \city{Edmonton},  \state{Alberta}, \country{Canada}}}

\affil[2]{\orgname{AI Redefined Inc}, \orgaddress{ \city{Montreal}, \state{Quebec}, \country{Canada}}}

\affil[3]{\orgname{JACOBB}, \orgaddress{ \city{Montreal}, \state{Quebec}, \country{Canada}}}

\affil[4]{\orgname{Thales Digital Solutions}, \orgaddress{ \city{Montreal}, \state{Quebec}, \country{Canada}}}

\affil[5]{\orgname{Alberta Machine Intelligence Institute}, \orgaddress{ \city{Edmonton},  \state{Alberta}, \country{Canada}}}


\abstract{
Recent advances in reinforcement learning (RL) and Human-in-the-Loop (HitL) learning have made human-AI collaboration easier for humans to team with AI agents. Leveraging human expertise and experience with AI in intelligent systems can be efficient and beneficial. Still, it is unclear to what extent human-AI collaboration will be successful, 
and how such teaming performs compared to humans or AI agents only. In this work, we show that learning from humans is effective and that human-AI collaboration outperforms human-controlled and fully autonomous AI agents in a complex simulation environment. In addition, we have developed a new simulator for critical infrastructure protection, focusing on a scenario where AI-powered drones and human teams collaborate to defend an airport against enemy drone attacks.
We develop a user interface to allow humans to assist AI agents effectively. 
We demonstrated that agents learn faster while learning from policy correction compared to learning from humans or agents. Furthermore, human-AI collaboration requires lower mental and temporal demands, reduces human effort, and yields higher performance than if humans directly controlled all agents. In conclusion, we show that humans can provide helpful advice to the RL agents, allowing them to improve learning in a multi-agent setting. 

}


\maketitle

\section{Introduction}\label{sec:intro}
Protecting critical infrastructure, such as an airport, against security threats is a complex, sensitive, and expensive task, leading to a history of exploring automated and autonomous solutions \cite{kewley2002computational}. However, fully automated and autonomous solutions in critical applications are not advisable due to the current limitations in technology maturity and trained operators. These might lead to poor performance, significant infrastructure damage, and increased risks of other collateral damages. Besides, training humans to utilize such solutions effectively remains a considerable challenge. On the other hand, continuous surveillance of such systems, quick assessment, and handling of potential threats would benefit from AI capabilities. 
In many cases, AI agents need assistance achieving full autonomy within a reasonable time frame due to the system's complexity or the scarcity of data \cite{gottipati2023hiking}. Another significant challenge is the AI agent\text{'s} ability to capture contextual understanding. For instance, consider an airport security scenario where an AI system affiliated with the airport authorities detects rapid movement on a surveillance camera or drone during nighttime. This system might classify the movement as an intruder, lacking the contextual nuance to recognize it as a routine patrol by the local police forces at the airport\text{'s} perimeter.

Humans generally possess domain expertise, experience, and contextual understanding in solving complex problems that are difficult for AI agents to learn or replicate. For example, considering the above example, a human operator might recognize the drone as a routine patrol based on the circumstances surrounding the drone\text{'s} presence and behavior. At the same time, the AI agent lacks the knowledge to respond appropriately.
Human decision-making becomes essential in safety-critical applications, where scenarios may be partially anticipated. Considering the value of human expertise, it is necessary to effectively leverage human knowledge and situational awareness in collaborative environments, especially for critical applications like defense or security. These applications are likely to benefit from systems that combine the strengths of both human operators and autonomous systems. This integration aims to decrease system costs and enhance task performance while maintaining meaningful human control in dangerous or critical operations. Such a hybrid approach is crucial to mitigate potential risks in these high-stakes environments \cite{santoni2018meaningful}.

Recently, reinforcement learning (RL) has successfully solved many complex decision-making problems, such as mastering the game of Go~\cite{silver2016mastering}, deploying super pressure balloons in the stratosphere~\cite{bellemare2020autonomous}, and generating synthetic drugs~\cite{gottipati2020learning, gottipatimaxbellman}. Although established domains like Atari and Mujoco serve as benchmarks for cutting-edge RL research~\cite{todorov2012mujoco, mnih2015human}, the introduction of simulators for complex domains facilitating human-AI collaboration has been less explored~\cite{schelble2020designing, o2022human}. However, a notable challenge in deep RL is its sample inefficiency~\cite{ibarz2021train}, requiring millions of interactions with the environment, making it difficult to adapt to real-world problems. To mitigate this, advice giving techniques such as demonstrations~\cite{hester2018deep, kharyalasp, mandlekar2022matters}, action-advice~\cite{torrey2013teaching, frazier2019improving, ilhan2021action}, preference~\cite{ibarz2018reward, palan2019learning, lee2021pebble} and reward shaping~\cite{ng1999policy, devlin2012dynamic, brys2015reinforcement, hu2020learning} have been used to guide RL agents to relevant parts of the state-space. However, most of this work has been restricted to game domains and advice by trained agents. A significant and relatively unexplored aspect concerns the potential improvement of human-agent collaboration through human demonstrations in complex, real-world environments. 
Furthermore, the current literature on human-agent collaboration reveals a noticeable scarcity of intelligent user interface design and integration for humans to provide effective advice. This scarcity frequently leads to misunderstandings between humans and AI agents, hindering the use of the human operator's expertise.

To address the challenges of complex real-world domains, we develop a novel simulator and user interface for the specific problem of the airport\text{'s} restricted zone protection system. The use case consists of a fleet of ally drones trying to protect restricted airspace against multiple intruding drones. Following recommendations from air defense domain experts, the simulator is designed to mimic the dynamics of a real-world scenario. This encompasses the drones' velocity, flight dynamics, the specifications of the ground radar sensor, the sensing payloads (radar and electro-optical), and the neutralization payloads embedded in the blue drones. Such real-world dynamics make the environment complex. The complexity of the environment means that a naive RL agent would require many environment interactions to learn an optimal policy. Given the cost and risk associated with these interactions in the specified domain, the trained agent needs to be sample-efficient. We demonstrate that learning from human or agent demonstrations can minimize the number of required environment interactions for the mentioned complex environment. 
Some research \cite{humann2019human, barnes2015designing, porat2016supervising} indicates that when one person oversees multiple agents in complex systems, the increased monitoring demand can negatively affect their workload and cognitive load — which can ultimately hinder performance. 

We demonstrate that better decision-making capabilities of the trained agents can reduce the human operator's workload and increase the performance of the human-agent team. The main goal of creating human-agent collaborations is to capitalize on the strengths of agents and humans while mitigating their weaknesses. For instance, intelligent agents excel in tasks such as analyzing vast data sets and making rapid decisions based on specific patterns, outperforming humans \cite{meyer1994deontic}. In contrast, humans exhibit superior decision-making abilities rooted in their moral values and contextual understanding, compared to agents \cite{van2021moral}. A characteristic feature of the specific defense domain use case is that operations are versatile, often highly unpredictable, and the ethical stakes can be extremely high. To maintain the exercise of authority and direction by humans, we also use human policy correction to correct the trained agent's policy. We show that online policy correction is the most effective form of advice to improve agent learning and achieve the best performance. In addition, we demonstrate that the cognitive workload of humans is lower while doing policy correction than that of humans controlling an untrained agent (drones in this domain). We use non-expert human and agent demonstrations to showcase the robustness of our approach to address the limited availability of human experts.

\medskip
\noindent \textbf{Contributions:} This article makes the following contributions:
\begin{enumerate}
    \item Introduces a novel multi-agent simulator for defense-specific airport protection use case modeling real-world dynamics with multiple ally and enemy drone agents.
    \item Uses state-of-the-art deep RL algorithms to train multiple agents inside the novel simulator. 
    \item Develops a user interface inside the simulator, which enables human operators to dynamically take control of single or multiple agents to produce in-context demonstrations, thus enabling human-agent collaboration.
    \item Demonstrates empirically that trained agent demonstrations or a mixture of human and agent demonstrations help the agent to learn faster. 
    \item Compares and evaluates multiple advice-giving techniques, i.e., learning from demonstration and policy correction. 
    \item Compares the human cognitive workload for various advice-giving techniques using a user study demonstrating that policy correction requires less effort than humans having full control over the agents. 
\end{enumerate}

\section{Related work} \label{sec:literature}

In deep RL, external knowledge from different sources makes RL agents sample-efficient \cite{argerich2020tutor4rl, bignold2021conceptual}, where this external knowledge can originate from humans or other agents. Early examples of using human expertise in decision-making relied on collecting and learning from demonstrations by experts using techniques such as imitation learning~\cite{schaal1999imitation}. However, these methods usually incur human costs in terms of mental and physical efforts, attention, and availability of experts \cite{biyik2022learning}. Although imitation learning methods are still popular, a significant amount of existing work uses human preference, feedback, policy shaping, and reward shaping as advice to better guide RL agents, rather than demonstrations from experts to guide RL agents better ~\cite{christiano2017deep,park2021surf,sumers2021learning, xue2023reinforcement}.

\textbf{Learning from demonstrations} \cite{11aamas-hat-taylor} has been a common approach to make  Deep RL sample-efficient. Hester et al.~\cite{hester2018deep} first leveraged demonstrations inside a deep Q-network (DQN) using a supervised loss function to account for deviations from expert demonstrations. Later, demonstrations were used to speed up training with DDPG in complex robotics tasks~\cite{nair2018overcoming,vecerik2017leveraging} by using specific techniques like behavior cloning loss and prioritized replay buffers. Goecks et al.~\cite{goecks2020integrating} provided a unified loss function by integrating loss function components from prior works. Existing literature \cite{knox2012reinforcement, warnell2018deep, arakawa2018dqn, 2017icml, arumugam2019deep} focuses on feedback or demonstration efficient algorithms where the authors primarily investigate several sampling and exploration strategies \cite{kim2013learning, schaul2015prioritized, andrychowicz2017hindsight, liang2021reward}, improve feedback efficiency using imitation learning \cite{ibarz2018reward} and unsupervised pre-training \cite{lee2021pebble}. 

Some existing work has used dueling double deep Q network \cite{hessel2018rainbow} (D3QN) to solve similar real-world complex tasks like security patrolling \cite{venugopal2021reinforcement}, path planning for an unmanned aerial vehicle (UAV) in a dynamic environment \cite{yan2020towards}, unmanned ground vehicle (UGV) control \cite{yuan2021centralised}, manufacturing \cite{oliff2020reinforcement}, vehicle-to-vehicle communication \cite{ji2023multi}, UAV autonomous aerial combat \cite{kong2020uav, jiang2022novel}. However, existing literature primarily focuses on improving autonomous agent performance, while our goal is to learn from humans and improve human-agent collaboration performance in multi-agent settings. Our main goal differs from those mentioned above, as we do not focus on sampling techniques or improving feedback efficiency through imitation learning. Our objective is to leverage human demonstrations in complex real-world defense scenarios to enhance the performance of human-AI collaboration. 

\textbf{Human-AI collaboration} is emerging as a critical field, integrating human and AI capabilities for diverse applications \cite{kamar2012combining, lasecki2012real, kamar2016directions}. Research has concentrated on enabling natural and efficient collaboration in human-AI systems, focusing on the communicative impact of shared actions \cite{liang2019implicit} and balancing performance gains with compatibility to human mental models \cite{bansal2019updates}. Effective human-AI collaboration in complex physical scenarios relies on seamless agent performance in simulations \cite{hoffman2004collaboration}. Notably, human-AI collaboration can expedite sequential manipulation tasks \cite{hayes2015effective} and enhance resource distribution~\cite{10.1145/3394287, herse2022optimising}. 
Autonomous AI agents, distinct from expert systems, can learn tasks without preprogramming \cite{phulera2017analytical}. Choudhury et al. \cite{choudhury2019utility} compare deep learning-based human models with structured \textquotedblleft theory of mind\textquotedblright models. Sandrini et al. \cite{sandrini2022learning} address the human-AI teaming planning and allocation problem using a minimum-time formulation. Jahanmahin et al. \cite{jahanmahin2022human} explore human-centered robot interaction in industrial settings, while Tambe et al. \cite{tambe2011security} did it in Stackelberg games settings. Our research examines the utility of human involvement in such settings, highlighting the need for shared autonomy, clear agent communication in real time in human-AI collaboration.

For tasks requiring human-agent collaboration, autonomous agent pilots are often paired with human counterparts to work together effectively \cite{tambe1998implementing, tambe1995intelligent}. Similarly, cognitive assistants are used to support astronauts during pivotal space missions, enhancing their decision-making capabilities \cite{van2009policy}. Another burgeoning area of research encompasses deploying human-AI collaborative teams in rescue operations \cite{hong2019investigating} and the evolution of semi-autonomous vehicles, where human operators and autonomous agents collaboratively navigate to destinations. Numerous existing studies \cite{hu2022autonomous, xin2022drl, cao2023autonomous} employ DQN or its variations to empower UAVs to autonomously formulate control commands and execute air combat tasks, responsive to the information derived from their environmental contexts. However, human involvement was missing or limited in all these prior works. Recently, Zhang et al. \cite{zhang2023uav} used D3QN with expert experience storage mechanism for decision-making in a simulated UAV Air Combat. They use expert experience with D3QN similar to ours and demonstrate effective use of training data and faster algorithm convergence. 
However, we use humans for providing demonstrations and policy correction, while existing works only use trained agent experiences. In addition, they also differ from our problem settings and layered architecture for the defined use-case. 


\textbf{Policy Correction in RL} is essential in safety-critical applications aiming to adjust strategies for optimal results \cite{liu2021policy, zhang2022saferl}. Off-policy correction, addressing policy divergence due to function approximation, bootstrapping, and off-policy errors, are often managed through importance sampling \cite{sutton1998reinforcement, munos2016q, owen2013monte}. Techniques such as inverse RL, behavior cloning, policy shaping, and constrained RL employ expert demonstrations, feedback, or constraints for policy guidance \cite{vecerik2017leveraging, kang2018policy, cederborg2015policy, knox2009interactively, satija2020constrained, marwan2021just}. Bai et al. propose a dynamic constraint set method for policy refinement using probability metrics \cite{bai2023picor}. 
In recent work, Zawalski et al. \cite{zawalski2022off} used importance sampling for off-policy correction in multi-agent reinforcement learning. In contrast to prior work, our focus shifts toward the impact of human correcting a trained agent's policy. We aim to assess the effects of human interventions on the performance of human-agent teaming compared to separate human and agent teams. 
Agents can learn desired behaviors by observing human experts, making demonstrations a valuable tool in policy correction \cite{celemin2019interactive}. In this work, we use humans to correct the policy of a trained agent in our airport security scenarios.

\section{Background} \label{sec:background}

In this section, we introduce the necessary background details on which our work is built. The subsequent sub-sections contain details related to RL, Deep Q Networks, and using demonstrations to guide Deep Q Networks.

\subsection{Reinforcement Learning (RL)}\label{sec:RL}
Reinforcement learning, multi-agent reinforcement learning, and other human-in-the-loop learning algorithms are often modelled as a Markov decision process (MDP), defined by a tuple $\langle\mathcal{S}, \mathcal{A}, T, \textit{R}, \mathcal{\gamma}\rangle$. Here, $\mathcal{S}$ represents the state space, and $\mathcal{A}$ denotes the action space. 
At each time-step $t$, an agent in state $s\in\mathcal{S}$ selects an action $a$ from action-set $\mathcal{A}$. The environment's transition probability, $T$, then determines the probability $p(s',r|s,a)$ of transitioning to state $s'$ and receiving reward $r$ given the current state $s$ and action $a$. The reward function $R: \mathcal{A} \times \mathcal{S} \rightarrow \mathbb{R}$ maps states and actions to scalar-valued rewards. The agent receives a reward $r_t$ after interacting with the environment. The agent's objective is to maximize the expected sum of discounted rewards, $G_{t}$, at any time-step, $t$:
\[G_{t} 	\doteq \sum_{k=0}^{\infty}\gamma^{k}r_{t+k+1} \]

\subsection{Deep Q-Network (DQN)}\label{sec:D3QN}
Deep Q Networks \cite{mnih2015human} use deep neural networks to estimate $Q$ values, updated via the Bellman equation. Further improvements were shown in Double DQN \cite{doubledqn} by addressing overestimation in DQN by decoupling action selection and evaluation. Specifically, Double DQN uses two Q-networks, each with different weights $\theta$ and $\theta^-$. The network with weights $\theta$ is used for action selection, while $\theta^-$ estimates the greedy policy\text{'s} value. The Q-value update in Double Q-Learning is given by:

\begin{equation} \label{eq:ddqn}
Q(s, a; \theta) = R_t + \gamma Q(s', \text{argmax}_{a'} Q(s', a'; \theta); \theta^-)
\end{equation}

\noindent Where $\theta$ and $\theta^-$ represent the parameters of the prediction network and the target network, respectively. Dueling DQN \cite{wang2016dueling} further advances DQN by introducing a dual network architecture, optimizing the Q network. It decomposes the Q-value into the state value function $V(s)$ and the action advantage function $A(s, a)$, allowing for more nuanced representation and improved decision-making in scenarios where the Q-value depends only on the state. The actual combination used in dueling DQN is as follows:

\begin{equation} \label{eq:duelling}
Q(s, a; \theta, \alpha, \beta) = V(s; \theta, \alpha) + \left( A(s, a; \theta, \beta) - \frac{1}{|A|} \sum_{a'} A(s, a'; \theta, \beta) \right)
\end{equation}

\noindent \noindent This ensures that for each state $s$, the advantages of all actions average to zero. Here, $ V(s) $ represents the value of state $s$, and $ A(s, a) $ represents the advantage of taking action  $a$ in state $s$. The parameters $\theta$  are shared across both functions, while  $\alpha$ and $\beta$ are unique to the state value and advantage functions. By integrating the strengths of both double DQN and dueling DQN architectures, we employ D3QN for agent learning in this work to achieve enhanced performance in complex environments involving high-dimensional state and action spaces.

\subsection{Deep Q learning from demonstration (DQfD)} \label{sub:DQfD} 
We used Deep Q learning from demonstration \cite{hester2018deep} for the discrete action scenario, combining human demonstrations with agent experiences for stability and efficiency. DQfD, integrated into the D3QN framework, accelerates policy optimization by leveraging expert experience. It employs an additional supervised loss function alongside Q-learning loss, ensuring that the agent prioritizes actions from expert demonstrations. In DQfD, the agent is pre-trained with demonstrations using a margin classification loss to mimic expert behavior closely. The margin classification loss is defined as below:

\begin{equation*}\label{eqn:margin}
J_E(Q)=\max _{a \in A}\left[Q(s, a)+l\left(a_E, a\right)\right]-Q\left(s, a_E\right)
\end{equation*}
where $a_E$ signifies the action executed by the demonstrator in state $s$. The margin function $l(a_E, a)$ is zero when $a = a_E$ and positive otherwise. The $n$-step return is used to propagate the values of the demonstrator's trajectory to preceding states. The $n$-step return is defined as:
\begin{equation*}
r_t+\gamma r_{t+1}+\ldots+\gamma^{n-1} r_{t+n-1}+\max _a \gamma^n Q\left(s_{t+n}, a\right)
\end{equation*}
The subsequent $n$-step loss, accounting for the $n$-step return, is denoted as $J_n(Q)$. An L2 regularization loss was also introduced to prevent over-fitting to the limited demonstration dataset. The comprehensive loss used for network updates is:
\begin{equation} \label{eqn:ddqn_demo}
J(Q)=J_{D Q}(Q)+\lambda_1 J_n(Q)+\lambda_2 J_E(Q)+\lambda_3 J_{L 2}(Q)
\end{equation}
\noindent where  $\lambda_1$, $\lambda_2$ and $\lambda_3$ parameters control the weighting between the losses. Hester et al. \cite{hester2018deep} employed DQfD in a single-agent RL context. In this work, we extend its application to a multi-agent RL setting, training all five ally drones based on demonstrations and the collective experiences of all agents. Further details are given in Sections \ref{sec:problem formulation} and \ref{ch:experiments}.

\section{Problem Formulation} \label{sec:problem formulation}

In this section, we describe the airport defense use-case and formulate the problem as an MDP. We also describe the user interface and relevant details about the Human-in-the-loop interactions. Details related to the system architecture of the environment are provided in Appendix \ref{app:hilt_user-interface}.

\subsection{Environment Design} \label{Environment_design}

\begin{figure}[h]
  \centering
  \includegraphics[clip,width=0.8\columnwidth,scale=1]{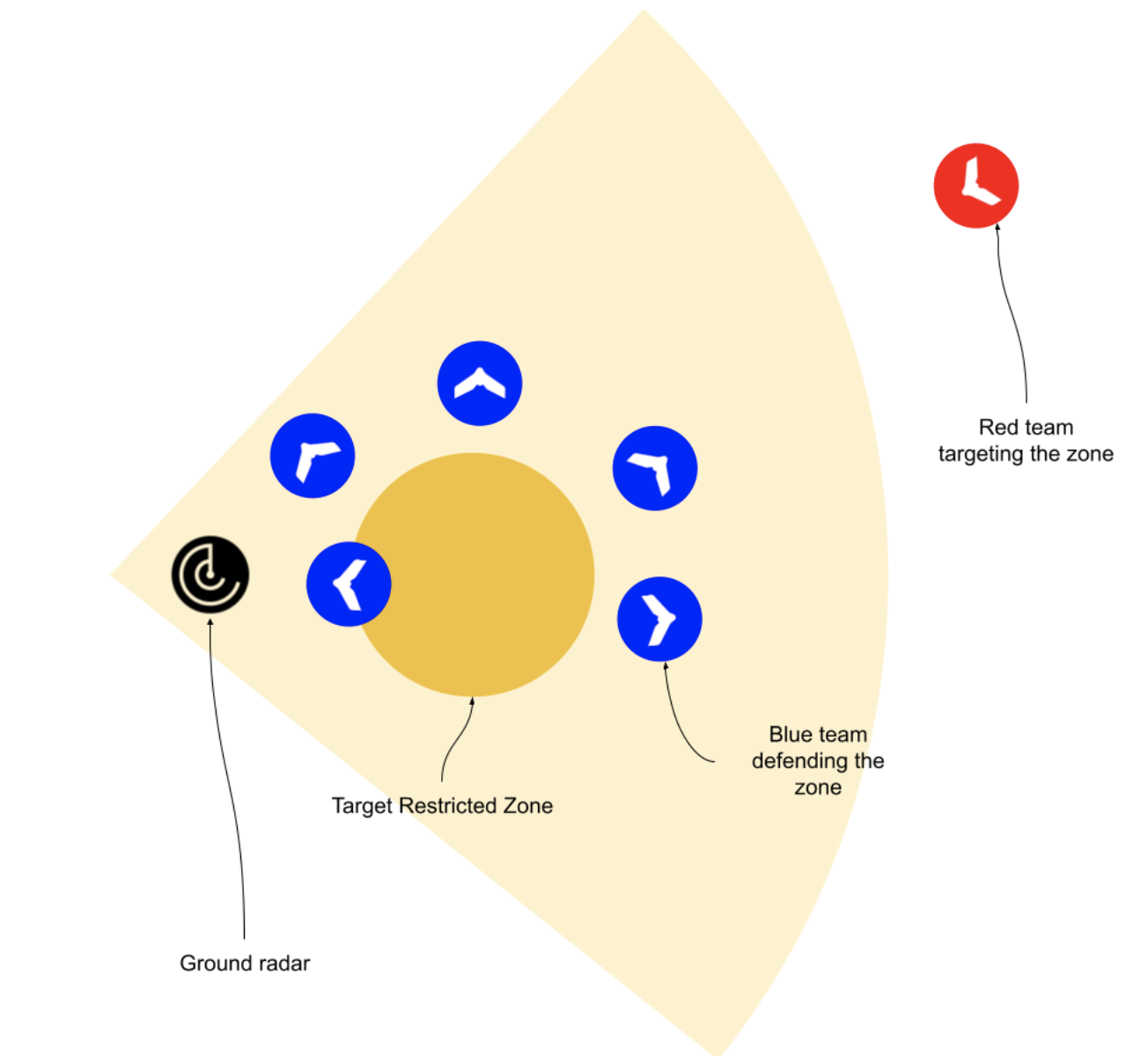}
  \caption{Environment: airport defence scenario} 
  \label{img:thunderblade}
\end{figure}

In this work, we introduce an airport defense simulator and explore the impact of human demonstration on this domain by formulating it as a multi-agent RL problem. The environment used for our problem formulation is shown in Figure~\ref{img:thunderblade}. In this stochastic environment, a team of ally drones collaborates to achieve the shared objective of securing the airport\text{'s} restricted zone from enemy intrusions. In contrast, the enemy drones plan attacks with knowledge about the defender\text{'s} strategy. The ally (blue) drones and humans aim to ensure airport security by working as a team to counter any threats posed by enemy (red) drone attacks. The ally (blue) drones and humans aim to ensure airport security by working as a team to counter any threats posed by enemy (red) drone attacks.

\noindent The blue drones can be autonomous, remotely piloted, or controlled through occasional human interventions. 
Each drone is equipped with a gimbal-mounted electro-optic sensor, allowing it to capture data that can be leveraged, e.g., for surveillance or threat level assessment. The blue team comprises five drones, a ground radar sensor, and a ground control station (GCS). Each ally drone also has several neutralization payloads (i.e., devices capable of neutralizing enemy drones when they are within a specific range). The red team comprises a single drone equipped with its radar sensor and a potentially hazardous payload. The goal of the blue team is to detect, localize, intercept, and neutralize the enemy drones before they reach the restricted zone of the airport.
Details of the ally drones team and enemy drone are given in Appendix \ref{app:team_description}.

The experimental platform is built around a simplified airspace simulator operating in 2D. Although simplified, several aspects have been modeled following real-world specifications based on feedback from domain experts, such as the detection capabilities of the drone sensors and the radar, as well as the dynamics of the fixed-wing drones. In this environment, the blue and red drones have a partially observable view of the environment. The detection and localization of the red drone provided by the radar and EO sensors embed noise and uncertainty.
As per the defense expert\text{'s} suggestions, we introduce errors into the system to simulate these real-world factors. Specifically, the radar detection probability is 95\%, and there is a 5\% probability of the radar failing to detect the red drone in 1 second. 
Moreover, the radar fails to detect the enemy drone if it is outside the radar range. The detection frequency is set to 1~Hz, and the maximum speed of the drones is $10$ meters per second. The range of the neutralization payloads embedded on the blue drones is set to $10$ meters. All these dynamics make the scenario complex and require the blue team to anticipate the trajectory of the red drone to neutralize it.

\subsection{MDP Formulation}
\label{sec:MDP formulation}

We model the above-mentioned problem as a Markov decision process as defined below:
\begin{enumerate}
    \item \textbf{State space:} The state space consists of the relative positions of 1) the red drone, 2) the blue drones, and 3) the restricted airspace over $3$ time steps. 
    In our multi-agent setting comprising five drones, each drone has a partial observation of the environment, where it lacks the capability to perceive the presence and actions of its peer drones directly. Each blue drone has an observation that includes 1) the relative distance to the red drone if detected along the x and y coordinates (in meters) and 2) the relative distance to the centre of the restricted zone along the x and y coordinates (in meters). To capture the context beyond the current drone position, we aggregate drone positions over three consecutive time steps, resulting in a state with $(2 + 2) \times 3 = 12$ features by stacking 3 consecutive time steps together.
    
    \item \textbf{Action space:} The action space consists of a single continuous action, rotation, which lies in the range $[-1,1]$. We discretize this into $2$ discrete actions: positive and negative rotation. The agent must choose between a positive or negative rotation at each time step. 
    
    \item \textbf{Reward function:} The blue drones receive a positive reward if they successfully neutralize the red drone and a negative reward if the red drone enters the designated target area.  
    
    \noindent The team's reward function is defined as follows:
     \begin{equation*}
         R(s) = \begin{cases} 
            +1 & \text{if any blue drone neutralizes the red drone}\\
            -1 & \text{if the red drone enters the restricted zone}
       \end{cases}
    \end{equation*}
 Additionally, at every time step, the blue drones receive a shaping reward, $R_{I}(s)$, proportional to their relative distance from the red drone in consecutive time steps
 \begin{equation*}
    R_{I}(s) \propto (d_{t-1}(b,r)-d_{t}(b,r))  
 \end{equation*}
 where $d_t(b,r)$ and $d_{t-1}(b,r)$ refer to the relative Euclidean distances between the blue drone and the red drone at time steps $t$ and $(t-1)$, respectively. We use potential-based reward shaping as the optimal policy is guaranteed to be invariant \cite{ng1999policy, devlin2012dynamic, brys2015reinforcement, hu2020learning}.
\end{enumerate}

\noindent We train the agents in multi-agent centralized training and decentralized execution (CTDE) settings where each agent has its own observation space as defined in the MDP formulation. Each agent has a similar reward function based on their current location and action.
We train these agents in parallel using Cogment~\cite{cogment}. Cogment is an open-source platform enabling training and operating various kinds of multi-agent RL and human-in-the-loop learning algorithms in a distributed way due to its microservice architecture. 

\subsection{User-interface for advice} \label{app:user_interface}

To foster effective human-agent collaboration within a simulated airport environment, we developed a user interface that comprehensively visualizes the airport and its nearby areas. This interface, developed using JavaScript and primarily leveraging the React front-end framework, serves as a nexus where autonomous ally drones and human operators collaboratively safeguard restricted zones, which are inherently vulnerable to potential adversarial drone incursions.

\begin{figure}[h]
  \centering
  \includegraphics[clip,width=0.9\columnwidth,scale=1]{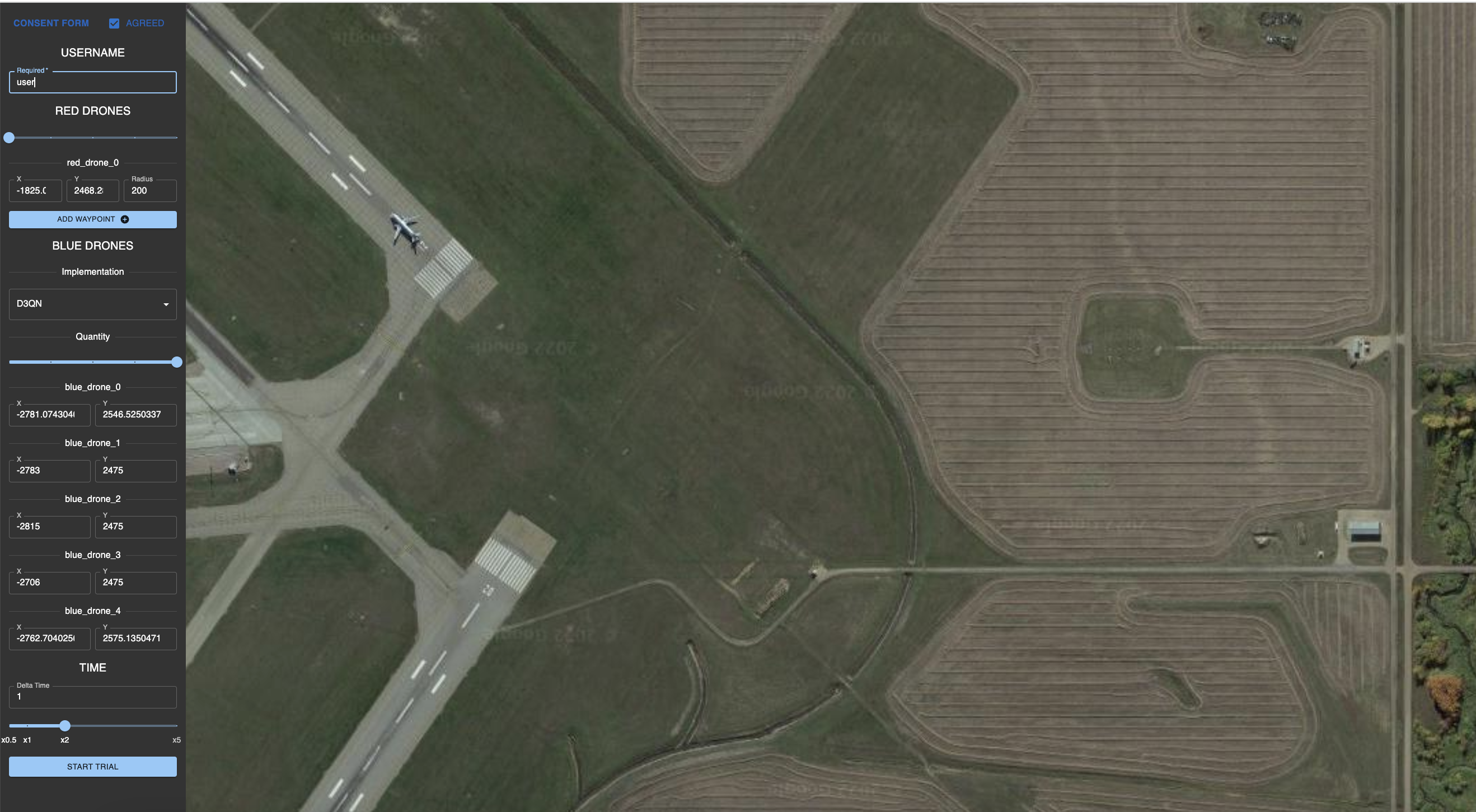}
  \caption{A trial configuration view of user interface.} 
  \label{img:interface_2}
\end{figure}

Our user interface has two distinct views: a trial configuration form and an interactive trial run time view. Figure \ref{img:interface_2} shows the starting view of the trial configuration form that empowers users with the flexibility to tailor various parameters. Specifically, users can modify the composition and positioning of the blue team drones, select their underlying AI algorithm, and decide on the potential involvement of a human operator. These options allow the user to operate and control the agent, where the agent is trained with the D3QN algorithm. Additionally, configurations extend to determining the number of adversarial red drones and charting their trajectory toward the restricted zone, with an added provision to stipulate the simulation\text{'s} update frequency.

\begin{figure}[h]
  \centering
  \includegraphics[clip,width=0.9\columnwidth,scale=1]{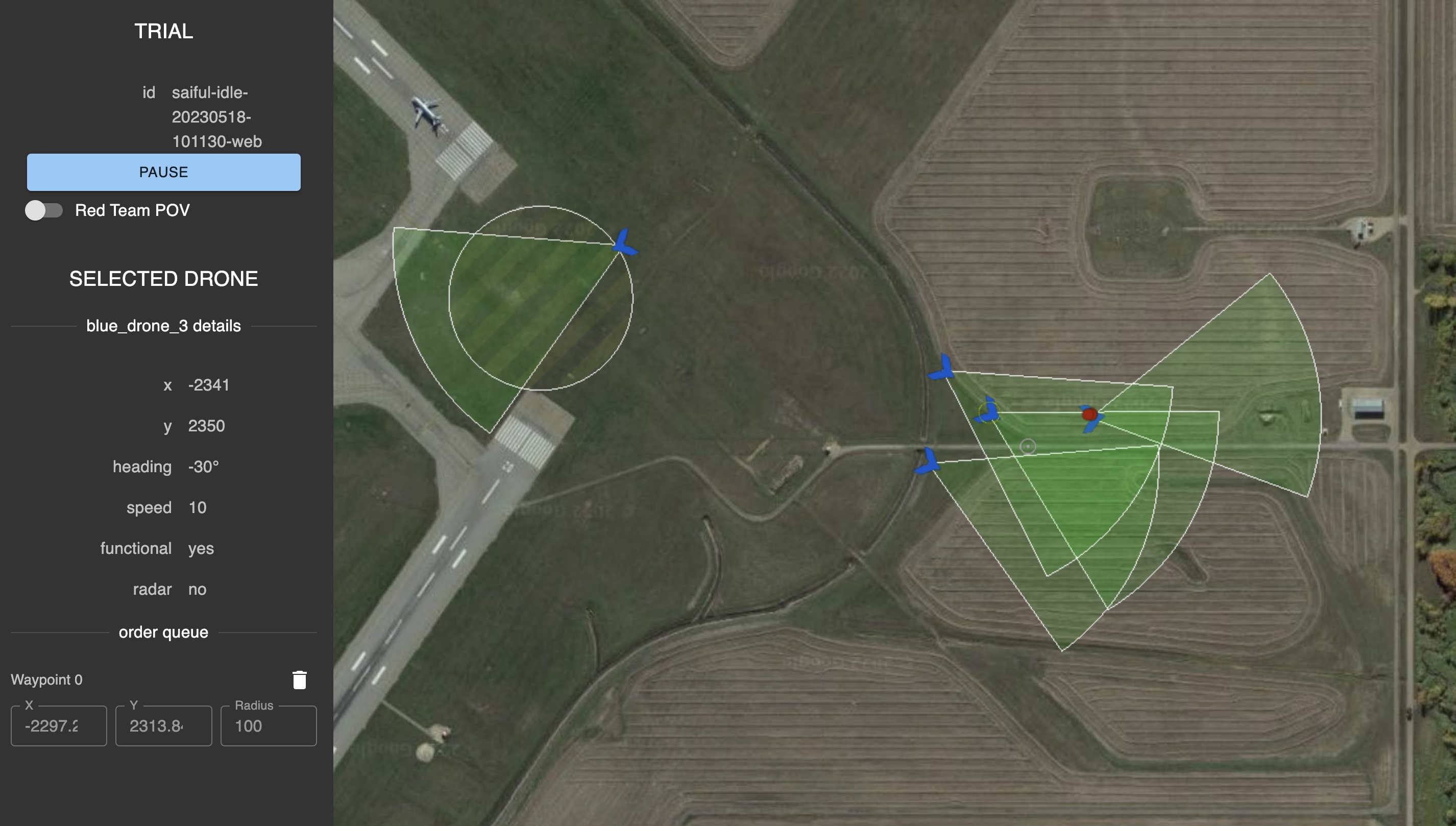}
  \caption{An interactive trial run time view user interface for human operators to control the agents}
  \label{img:interface_1}
\end{figure}

Transitioning to the trial run time interface, as depicted in Figure \ref{img:interface_1}, presents an aerial perspective of the environment. This view delineates the drones, their respective detection radii, ground radar, and any identified red drones. Enhanced interactivity is facilitated by enabling users to select individual drones, assign or eliminate waypoints by interacting directly with the map, and manage the simulation's progression by pausing or resuming it. To control the ally drones or modify the drone trajectories, humans can define the waypoints for each drone separately. To \textit{add} a waypoint, a human can select the drone he wants to control using a right mouse click and choose the point where the agent needs to go as a next step using a left mouse click. Users can add as many waypoints as they want to control the drone.
Similarly, humans can delete waypoints from the left side of the panel using the delete button shown on the bottom left side. To \textit{delete} specific drone\text{'s} waypoint operators need to select the drone using the mouse first; operators can also delete any waypoints of the selected drone. Furthermore, users can seamlessly pan and zoom within the map for a more detailed inspection. This interface aims to provide a platform for managing mission-critical data and facilitating user interaction and control over multiple agents. Users can act as  \textit{operators}, providing demonstrations to guide agents or intervene to correct the policy of a trained agent.

\section{Experiments}\label{ch:experiments}

In this section, we report the environment configuration, experimental settings, and performance metrics used in our experiments.

\subsection{Environment Configuration}\label{ec:environment}
\begin{figure}[h]
  \centering
  \includegraphics[clip,width=0.9\columnwidth,scale=1]{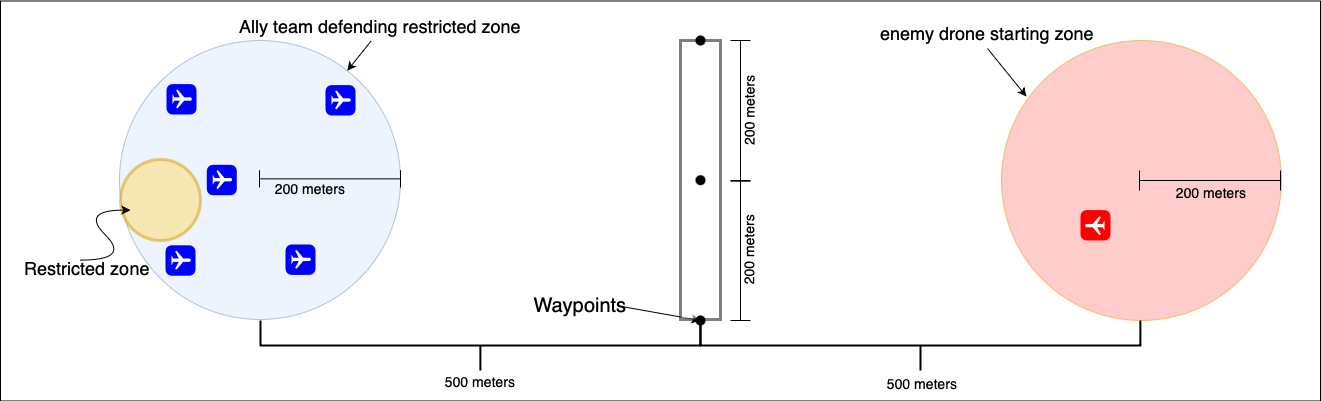}
  \caption{Three waypoint scenario (Simple scenario)
  } 
  \label{img:HAT_scenario_1}
\end{figure}

We used two distinct environment configurations for our experiments based on recommendations from defense experts:

\noindent \textbf{Three Waypoint Scenario (Simple Scenario):} 
In this setup, the starting positions of the five blue drones are determined within a circular region of a 200-meter radius near the restricted zone, as shown in Figure \ref{img:HAT_scenario_1}. The red drone starts from a similar circular region, positioned 1,000 meters to the right of the restricted area. The starting positions of the drones are selected randomly at the beginning of every episode. 
We added three fixed waypoints between the starting positions of the blue and red drones. The Euclidean distance between two waypoints is 200 meters. These waypoints guide the red drone towards the restricted space within the time limits. 

\begin{figure}[h]
  \centering
  \includegraphics[clip,width=0.9\columnwidth,scale=1]{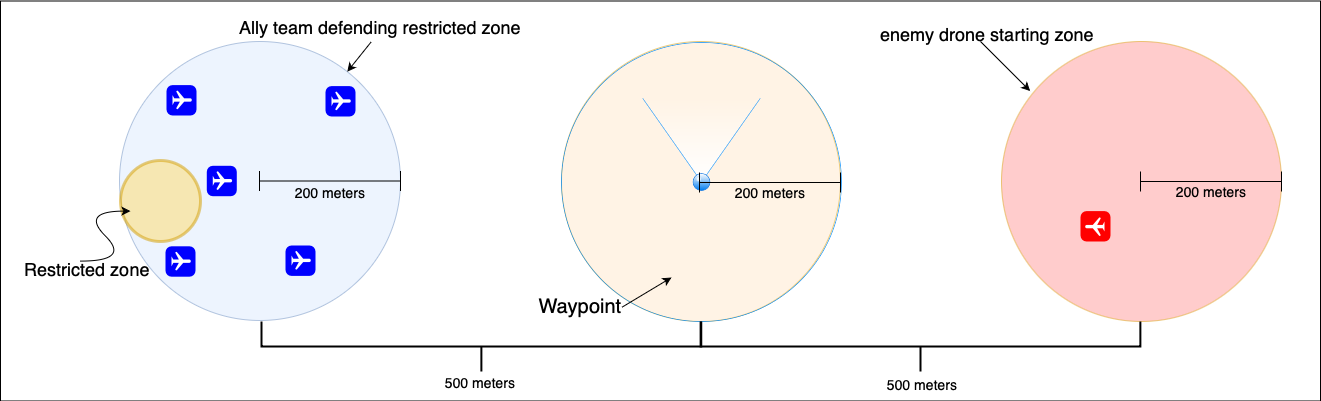}
  \caption{Continuous waypoint scenario (Complex scenario).
  } 
  \label{img:HAT_scenario_2}
\end{figure}

\noindent \textbf{Continuous Waypoint Scenario (Complex Scenario):} 
For this scenario, the blue drones' starting positions mirror those in the three waypoint scenario. However, the setup introduces only one random waypoint selected from a 200-meter radius circle between the initial positions of the blue and red drones, as shown in Figure \ref{img:HAT_scenario_2}. All the starting points for blue drones, red drone, and waypoints are selected randomly at the beginning of every episode. The random waypoint increases the uncertainty with respect to the location and makes the problem more complex. This waypoint is chosen strategically to prevent collisions between the blue and red drones.

\subsection{Performance Metric}
We evaluate our results across $2$ dimensions: (1) task success and (2) human effort while giving advice. We use team performance across the first dimension and cognitive workload measures based on NASA Task Load Index \cite{hart2006nasa, richards2020measure} questionnaires along the second dimension.

\noindent\textbf{Team Performance:}
We evaluated the performance of the trained agent using the \textit{success rate} as the performance metric. The success rate is the percentage of times the blue team wins overall evaluation trials. 
This metric was chosen because it provides a clear and intuitive measure of the agents' ability to defeat the red drone and is directly proportional to the average reward. 
We execute thirty evaluation episodes per hundred training episodes to compute the success rate. During the evaluation episodes, agents execute the learned policy without exploration.
Our learning curves show the performance metric reported across the evaluation episodes. 

\noindent\textbf{Cognitive Workload:}
The cognitive workload measures are assessed using the NASA Task Load Index ~\cite{hart2006nasa, richards2020measure}. This index evaluates six types of workload: mental demand, physical demand, temporal demand, performance, effort, and frustration. Each participant selects a score ranging from 1 to 21 using a 21-point slider for all six workload types.

\subsection{Demonstration Collection Methodology}
To collect demonstrations, a human teacher either controls the agent prior to training or a partially trained agent trained in the same environment with randomized starting position of the drones. We collected demonstrations in three ways: (1) the trained agent records the demonstration using agent\text{'s} sensors in the agent buffer. No mapping is required because the associated state-action pairs are captured directly from the trained agent's sensors. For our experiments, we use a fully trained D3QN agent to generate $2,500$ agent demonstrations. These demonstrations are referred to as \textbf{agent demonstrations} in our experiments.
(2) a human teacher demonstrates the task using our developed user interface, and the demonstration is collected in a database using a Cogment trial datastore and converted with embodiment mapping later for use. These demonstrations are referred to as \textbf{human demonstrations} in our experiments. 
(3) A human teacher controls a trained agent through the user interface and corrects its policy, thereby providing demonstrations, also referred to as\textbf{ policy-corrected demonstrations}. 
We engage eleven individuals for collecting demonstrations, collectively gathering 500 human demonstrations, with each participant completing at least 30 episodes. Similarly, we collected $500$ human demonstrations while correcting the policy of a trained D3QN agent. All the human demonstration data collection and human user study were done under the approval of the University of Alberta Research Ethics Board (REB number: Pro00107555). 

\subsection{Experimental Setup} \label{sec:exp_pro}

To establish a foundational benchmark and validate the functionality of the simulated environment, we implemented a heuristic-based decision-making algorithm tailored for drone operations. Our heuristic agents prioritize minimizing the distance to the enemy drone and either intercept their trajectory or follow their tracks for strategic positioning. In the subsequent sections, we use \textquotedblleft HA\textquotedblright{} to denote a heuristic-based agent. The \textquotedblleft HM\textquotedblright{} refers to the average winning percentage of human demonstrators who demonstrated through the designed interface.
The winning percentage for average human demonstration is $62\%$, with a standard deviation of $17\%$ in the simple scenario, and $64\%$, with a standard deviation of $7\%$ for the continuous complex scenario. We only considered demonstrations where either the agent or human won the game to account for good quality demonstrations. 

\begin{table}[h!]
\centering
    \begin{tabular}{ |c|c|c| } 
    \hline
    Algorithm abbreviation & Human Demo &  Agent Demo\\ [0.5ex] 
    \hline
        D3QN-0-0  & 0 & 0 \\ 
        D3QN-0-2500 & 0 & 2,500\\ 
        D3QN-500-2000 & 500 & 2,000\\
        D3QN$_{HM}$-500-0 & 500 & 0\\ 
        D3QN$_{PC}$-500-0 & 500 & 0\\
        D3QN-0-500 & 0 & 500\\
    \hline
    \end{tabular}
    \caption{\label{info-table} Algorithm abbreviation with the number of human and agent demonstrations for each algorithm. Here, the subscripts HM represent human demonstrations, and PC represents policy-corrected demonstrations.}
\end{table}

This study incorporated an autonomous agent programmed to learn and defend the airport from enemy attacks using DRL. Leveraging the capabilities of Q-value-centric DRL, we aim to achieve rapid policy convergence \cite{sutton1998reinforcement, kim2018learning, venugopal2021reinforcement}, thereby enhancing the feasibility of its application in more expansive systems. Our baseline was the D3QN algorithm, training from scratch without additional guidance, denoted as D3QN-0-0.  
We adopt deep Q learning from demonstration \cite{hester2018deep} with human experience replay buffer and agent experience reply buffer to leverage demonstrations inside D3QN from either agent, human or from a mix of human and agent demonstrations\footnote{Agent demonstrations refer to demonstrations generated by a D3QN agent that was trained to an average success rate of $85\%$ +/- $10\%$ in our airport security scenario.}. We use D3QN-0-2500 and D3QN-0-500 to represent D3QN agents trained with additional agent demonstrations of 2,500 and 500, respectively. D3QN-500-2000 denotes D3QN agents trained with a mix of 500 human demonstrations and 2,000 agent demonstrations. Table \ref{info-table} displays the proportions of human demonstrations and agent demonstrations used by each algorithm. In addition, we use D3QN$_{HM}$-500-0 and D3QN$_{PC}$-500-0 to represent D3QN agents trained with human and policy correction demonstrations, respectively. For experiments with D3QN-0-2500, D3QN-0-500, D3QN-500-2000, D3QN$_{HM}$-500-0 and D3QN$_{PC}$-500-0, we sampled 30\% demonstration samples from human or agent demonstration replay memory and 70\% from agent replay memory consisting of past agent experiences. 

After testing our models with various proportions of demonstration data, we determined that the learning algorithm\text{'s} performance was not significantly affected by the agent experience and demonstrations ratio, detailed as in Section \ref{sec:demo_portion}. Hence, we set these proportions to $70\%$ agent experience and $30\%$ experience from demonstrations for the experiments. To update the network, the training algorithm sampled mini-batches from the demonstration data and applied the double Q-learning loss and the n-step double Q-learning loss described in Section~\ref{sub:DQfD}. 
We did not use any pre-training for the reported experiments for fair comparison. The Q-function of D3QN is updated by equation~\ref{eq:duelling}; D3QN-0-2500, D3QN-0-500, D3QN-500-2000, D3QN$_{HM}$-500-0 and D3QN$_{PC}$-500-0 are updated using equation~\ref{eqn:ddqn_demo}. The expert margin in equation~\ref{eqn:margin} was set to $M=0.8$ in alignment with prior work \cite{hester2018deep}. 

All the reported experimental results are averaged over $5$ runs with different seed values, and standard deviations are reported.
The results in Figure~\ref{img:performance} and Figure \ref{img:performance_2} report the mean and standard deviation over $5$ runs, with the y-axis indicating the winning percentage and the x-axis denoting the number of training episodes. To determine significant differences between various baselines and our method, we employed the Mann-Whitney U test (also known as the Wilcoxon rank-sum test) \cite{mcknight2010mann}. This non-parametric statistical test was chosen for its robustness against deviations from normality assumptions, unlike other tests such as the t-test and ranked t-test. We observed that the normality assumption was violated in certain results, making the Mann-Whitney U test the preferred choice over the other tests. Hyper-parameter tuning was done using grid search to identify the best parameters for the algorithms, which were then consistently applied across all experiments in both scenarios as detailed in Appendix \ref{app:hyper-params}.

\section{Experimental Results}

Experiments were conducted in two scenarios (described in Section \ref{ec:environment}) to study the impact of different kinds of teaming.
We aim to investigate the following research questions:
\begin{enumerate}[label={RQ}{{\arabic*:}},leftmargin=1cm]
    \item How well does a trained RL agent perform in this environment? 
    \item Does agent and/or human demonstration help the RL agent learn more efficiently? 
    \item Does human policy correction help agents learn more efficiently, and how does this compare to learning with demonstration advice?
    \item How do humans experience this process? 
\end{enumerate}

\bigskip
\noindent We start by discussing the outcomes of the simple scenario, followed by an analysis of the results of the complex scenario. To evaluate the effectiveness of the agent and human demonstrations, we compare D3QN-0-2500, D3QN-0-500, and D3QN-500-2000 with HA, HM, and D3QN-0-0 as baselines.

\begin{figure}[h]
  \centering
  \includegraphics[clip,width=0.9\columnwidth,scale=1]
    {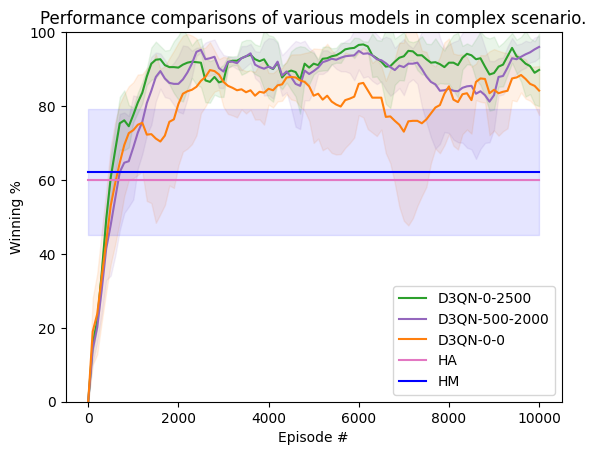}
  \caption{The success rate comparison for D3QN-0-0, D3QN-0-2500, D3QN-500-2000, HA, and HM performance in a simple scenario. Here, the number in the suffix represents the number of demonstrations from humans and the number of demonstrations from an agent. HA and HM represents heuristic based agent and the average winning percentage of human demonstrators, respectively.}
  \label{img:performance}
\end{figure}

To answer RQ1, we trained D3QN-0-0 on the simple scenario and reported its success rate in Figure~\ref{img:performance} based on 30 evaluation episodes per 100 training episodes. The D3QN-0-0 agents reach a success rate of approximately $90\%$ in $3,500$ episodes. The trained agent outperforms the baseline HA and HM, which have a success rate of $60\%$ and $63\%$, respectively.
D3QN-0-0 agents outperform both HA and HM. 

\begin{figure}[h]
  \centering
  \includegraphics[clip,width=0.9\columnwidth,scale=1]{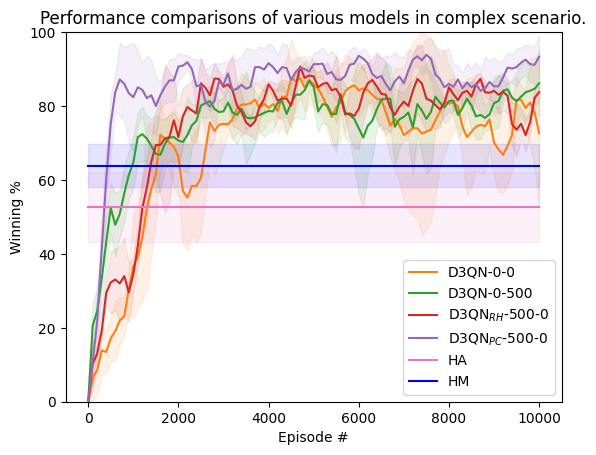}
  \caption{The success rate comparison for D3QN-0-0, D3QN-0-500, D3QN$_{HM}$-500-0, and D3QN$_{PC}$-500-0, HA, and HM performance in a complex scenario. Here, the number in the suffix represents the number of demonstrations from humans and the number of demonstrations from an agent. 
  }
  \label{img:performance_2}
\end{figure}

To answer RQ2, we see that D3QN-0-2500 reaches a success rate of more than $90\%$ in $1,600$ episodes, outperforming D3QN-0-0, as shown in Figure~\ref{img:performance}. In examining the levels of winning percentage between D3QN-0-0 and D3QN-0-2500, the Mann-Whitney U test results in Table \ref{tab:p-val-res-table} show a significant difference between them. At the end of learning, both algorithms converge to the same final performance (around $90\%$). This supports our claim that agent demonstrations make the RL agent more sample efficient in our environment, consistent with existing results in the literature~\cite{hester2018deep,nair2018overcoming}. 

We also trained the learning agent with a mix of agent and actual human demonstrations, denoted D3QN-500-2000, as shown in Figure~\ref{img:performance}. We sampled an equal proportion of human and trained agent demonstrations in every mini-batch used for training. We used a mix of both types of demonstrations due to the lack of human demonstrations collected in our user study. The Mann-Whitney U test shows no significant learning improvement when compared to D3QN-0-2500 in a simple scenario; however, the performance is still statistically significant than the baseline D3QN-0-0, shown in Table \ref{tab:p-val-res-table}.

\begin{table}[!h]
\centering
    \begin{tabular}{ |c|c|c|c|c| } 
    \hline
    Algorithm & P-value & U-statistic &  Effect size & Significance\\ [0.5ex] 
    \hline\hline
        D3QN-0-2500 vs.\ D3QN-0-0                 & 0.0001 & 134.5 & 0.832 & Yes \\
        D3QN-500-2000 vs.\ D3QN-0-0               & 0.0000 & 2050.5 & 0.435 & Yes \\
        D3QN-500-2000 vs.\ D3QN-0-2500            & 0.2790 & 687.0 & 0.141 & No \\
        D3QN-0-500 vs.\ D3QN-0-0                  & 0.0007 & 86.0 & 0.61 & Yes \\ 
        D3QN$_{HM}$-500-0 vs.\ D3QN-0-0           & 0.0061 & 111.0 & 0.497 & Yes \\ 
        D3QN$_{HM}$-500-0 vs.\ D3QN-0-500         & 0.5712 & 197.5 & 0.104 & No \\ 
        D3QN$_{PC}$-500-0 vs.\ D3QN-0-0           & 0.0001 & 1.0 & 0.0 & Yes \\ 
        D3QN$_{PC}$-500-0 vs.\ D3QN-0-500         & 0.0001 & 7.0 & 0.968 & Yes \\ 
        D3QN$_{PC}$-500-0 vs.\ D3QN$_{HM}$-500-0  & 0.0002 & 74.5 & 0.662 & Yes \\
    \hline
    \end{tabular}
    \caption{\label{tab:p-val-res-table}Comparison of different algorithm significance using the Mann-Whitney U test. Here, subscripts PC and HM represent policy corrected demonstration and human demonstrations, respectively.}
\end{table}

Similarly, for our complex scenario, we trained D3QN-0-0 and plotted its success rate in Figure~\ref{img:performance_2}. The agent reaches a success rate of $80\%$ in $3,500$ episodes, which is $10\%$ less than the D3QN-0-0 performance in a simple scenario. The trained D3QN-0-0 agent outperforms the baseline HA and HM, with a success rate of $53\%$ and $64\%$, respectively. With the complex scenario, the performance of HM remains the same or improves marginally, whereas the performance of HA and D3QN-0-0 drops significantly. This suggests that humans improve their performance playing more times and outperform HA in more complex environments that require greater generalization capabilities. In conclusion, despite the inherent randomness and complexity of the scenario that affects performance, our trained D3QN-0-0 agent shows superior performance compared to both Ha and HM, supporting our earlier finding regarding RQ1.

Figure~\ref{img:performance_2} presents the learning curves for D3QN-0-0, D3QN-0-500, D3QN$_{HM}$-500-0, and D3QN$_{PC}$-500-0 in our complex scenario, allowing a comparison of their performances. In particular, D3QN-0-500 shows a higher success rate compared to D3QN-0-0, as evidenced by the results of a Mann-Whitney U test (U=86.0, p=0.00075, effect size=0.61), which confirms significant performance differences between these two models. Additionally, D3QN$_{HM}$-500-0 shows superior performance in neutralizing enemy drones when compared to D3QN-0-0, as detailed in Figure \ref{img:performance_2} and shown in Table \ref{tab:p-val-res-table}. However, the performance gap between the D3QN$_{HM}$-500-0 approach and D3QN-0-500 is not statistically significant in our complex scenario. These findings are consistent with our conclusions from the simple scenario and answer our RQ2.

To address RQ3, we use policy-corrected demonstrations to train our agent, denoted as D3QN$_{PC}$-500-0. We find a significant performance increase in D3QN$_{PC}$-500-0 compared to all our baseline methods and learning from agent and human demonstrations. D3QN$_{PC}$-500-0 achieves a success rate of more than $80\%$ in $1,000$ episodes, while D3QN-0-0 takes $5,000$ or more episodes. The results of the Mann-Whitney U test show significant differences in performance between D3QN$_{PC}$-500-0 and D3QN-0-0, and also between D3QN$_{PC}$-500-0 and D3QN-0-500 (see Table \ref{tab:p-val-res-table}). Similarly, the Mann-Whitney U test between D3QN$_{HM}$-500-0 and D3QN$_{PC}$-500-0 ($U=74.5$, $p=0.00025$, effect size=0.662) indicates significant differences between them. This supports our claim that human demonstrations make the RL agent more sample efficient in our complex environment, as seen from the success of D3QN$_{PC}$-500-0 and D3QN$_{HM}$-500-0, consistent with previous results from the simple scenario. In addition, the policy correction approach shows a 10\% performance improvement over D3QN-0-0 and demonstrates the effectiveness of human-AI collaboration in our proposed scenarios, which answers RQ3 affirmatively.

\section{User Study Design} \label{sec:user_study}

To investigate whether human demonstrators can make RL agents more sample efficient, we have designed a human subject study where users can provide advice in two ways: (1) by providing full demonstrations, referred to as human demonstrations; and (2) by providing partial demonstrations and intervening when necessary, referred as policy-corrected demonstrations. Furthermore, to study the trade-off between human costs such as metal demand, physical demand, etc., and agent performance, this study contained survey questionnaires based on the NASA-TLX load index \cite{hart2006nasa, richards2020measure} and a set of demographic questions.  
We hosted our developed system on Amazon AWS to conduct
our user study and advertised the study on mailing lists of
graduate and undergraduate computing science students at the University of Alberta and industry partner organizations. The participants undertook a control task that involved controlling agents to neutralize the enemy in our simulated environment with randomly set environment configurations. Our user study was conducted using a structured approach with the following sequential steps.
\begin{enumerate}
    \item Read the task details and digitally sign a consent form.  
    \item Watch videos on how to use the web client and the user interface. In addition, watch videos of trained agents, controlling agent steps (setting up new waypoints or deleting the existing ones), providing a full demonstration, or providing a policy-corrected demonstration to teach the agents. This step is meant to teach the task to the participants.
    \item Round one: provide demonstrations for 30--40 episodes, controlling all the ally drones.
    \item Complete the NASA-TLX questionnaires based on their experience during the round one task.
    \item Round two: provide policy corrections for trained agents for 30--40 episodes.
    \item Complete the NASA-TLX questionnaires based on their experience during the round two task.
    \item Complete the demographics questionnaire. 
\end{enumerate}

\noindent In the initial phase of our study, we collected human demonstrations from Round 1, focusing on our simple scenario. Later, we expanded the study to include a complete user study comprising Rounds 1 and 2 in our complex scenario. We used the same set of participants for both rounds, while these rounds were conducted separately. Participants spent approximately 30--40 minutes on each round to complete the experiments.
To evaluate team-wise costs, such as mental, physical, and temporal demand, effort, etc., associated with human involvement, we calculate cognitive workload based on NASA-TLX questionnaires. In addition, demographic information collected included gender, age, current job/position, level of defense experience, level of drone experience, level of simulated drone control, and level of gaming experience.

\subsection*{Measurement of Cognitive Load}
To measure cognitive load, we conducted a user study described in Section \ref{sec:user_study} and used the NASA TLX survey questionnaires detailed in Appendix \ref{app:nasa_tlx}.
In our user study, 30 people consented to participate, and 11 both provided demonstration and completed the survey. The average age was 26.02 (SD of 6.42), ranging from 18 to 50.  
Analysis of the responses to the questionnaire indicated that 90\% of the participants reported being familiar with AI and games. In contrast, 80\% of the participants reported no prior familiarity with drone control or defense strategies.

\begin{figure}[h]
  \centering
  \includegraphics[clip,width=0.9\columnwidth,scale=1]
  {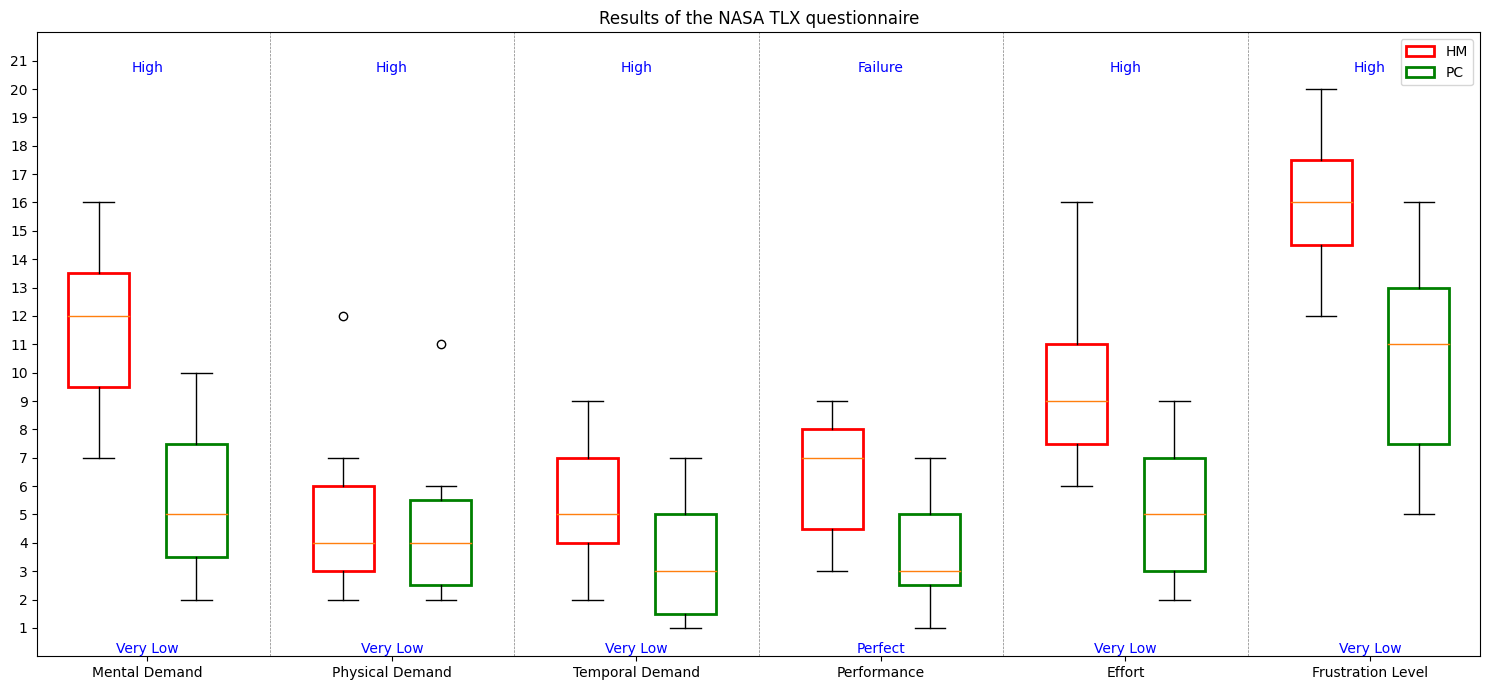}
  \caption{Results of the NASA TLX questionnaire. Here, PC represent policy correction by human and HM represents human demonstrations.}
  \label{img:nasa_tlx}
\end{figure}

\noindent As presented in Figure \ref{img:nasa_tlx}, participants reported much lower mental demand, temporal demand, and effort during policy correction denoted as \textquotedblleft PC\textquotedblright{} compared to human demonstrations denoted as \textquotedblleft HM\textquotedblright{}. Similarly, participants achieved much higher performance when they corrected the policies than when they fully controlled the agents (i.e., provided demonstrations). To evaluate the statistical significance of these observations, a Mann-Whitney U test was performed, comparing the cognitive load between PC and HM. As detailed in Table \ref{tab:nasa_mann}, the results show significant differences between PC and HM across all dimensions of the NASA Task Load Index (NASA-TLX), except for physical demand. These findings suggest that human operators experience more mental demands and frustration when fully controlling drones, especially in our complex simulator. In contrast, PC, a human-AI collaboration approach, requires less effort and yields superior performance outcomes. This addresses our fourth research question, RQ4, regarding the human experience in this process.

\begin{table}[h]
\centering
\begin{tabular}{lllll}
\hline
\textbf{Measure} & \textbf{P-Value} & \textbf{U-Statistic} & \textbf{Correlation} & \textbf{Significance} \\ \hline
Mental Demand    & 0.0006         & 113.0         & 0.8677                       & Yes                \\
Physical Demand  & 0.5711         & 69.5         & 0.1487                       & No                \\
Temporal Demand  & 0.0430         & 91.5         & 0.5123                       & Yes                \\
Performance      & 0.0098         & 100.0         & 0.6528                      & Yes                \\
Effort           & 0.0026         & 106.5        & 0.7603                       & Yes                \\
Frustration      & 0.0014         & 109.5         & 0.8099                       & Yes                \\ \hline
\end{tabular}
\caption{Mann-Whitney U Test for NASA-TLX between PC and HM, here correlation represents Rank-Biserial Correlation.}
\label{tab:nasa_mann}
\end{table}

\section{Ablation and Analysis}
We also conducted several ablation studies to examine the effect of proportions, quantity, and diversity of demonstration data.

\subsection{Effect of Demonstration Proportions} \label{sec:demo_portion}
\begin{figure}[h]
  \centering
  \includegraphics[clip,width=0.9\columnwidth,scale=1]{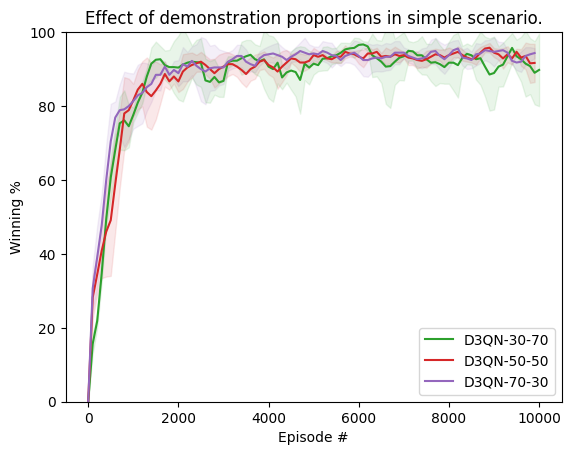}
  \caption{The success rate comparison for D3QN trained with 2,500 agent demonstrations(D3QN-0-2500) with various  proportions in simple scenario. Here the suffix represents the proportion of demonstrations from experience reply buffer and agent own experience, respectively.}
  \label{img:demo_percentage}
\end{figure}

We evaluated the training models by selecting varying proportions of demonstrations from experience reply memory and the agent\text{'s} own training experience from the environment in our simple scenario. Specifically, we combined demonstration samples from agent demonstration experience replay memory and 70\% from agent replay memory consisting of past agent experiences in ratios of 30:70, 50:50, and 70:30 percent, respectively, as shown in Figure \ref{img:HAT_scenario_1}. From our experiment, we concluded that trained agent demonstration and agent training experience from the environment proportions did not significantly impact the learning performance of D3QN-0-2500, as seen in Figure \ref{img:demo_percentage}. In addition, we conducted the Mann-Whitney U test, which yields a U-Statistic of 5556.5 with a p-value of 0.2197 and effect size of 0.340, indicating that there is no significant difference between D3QN-30-70 and D3QN-50-50 performance. Similarly, D3QN-50-50 and D3QN-70-30 show no significant difference with a U-Statistic of 5754.5, a p-value of 0.0653, and an effect size of 0.530.

\subsection{Effect of Demonstration Quantity} \label{sec:demo_quant}
\begin{figure}[h]
  \centering
  \includegraphics[clip,width=0.9\columnwidth,scale=1]
  {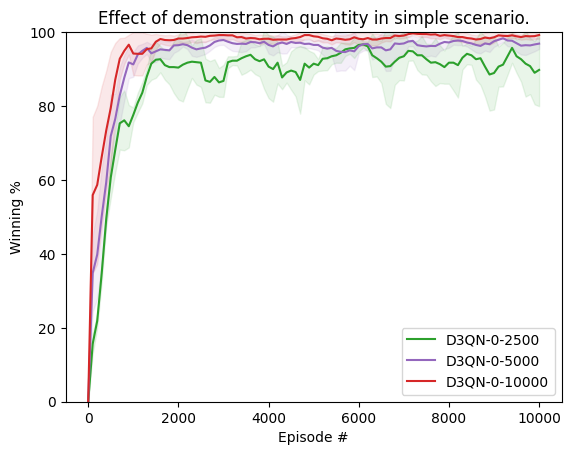}
  \caption{The success rate comparison for D3QN with agent demonstrations with various sizes of demonstration data in our simple scenario. Here, the suffix 2,500, 5,000, and 10,000 represents the total amount of agent demonstration in the experience reply memory.}
  \label{img:demo_amount}
\end{figure}

We also evaluated our approach in the simple scenario, as shown in Figure \ref{img:HAT_scenario_1} by increasing the number of demonstrations in the expert experience reply buffer. In all experiments, we use 30\% demonstration proportion from the expert reply buffer and 70\% from the agent experience reply buffer. We observed a significant improvement in agent\text{'s} performance with the increased number of demonstrations, as shown in Figure \ref{img:demo_amount}. 

\noindent The green curve represents the D3QN with demonstration, where data in the buffer was 2,500. The purple and red curves represent the D3QN guided by the demonstration, where data in the expert experience reply buffer was 5,000 and 10,000, respectively. A Mann-Whitney U statistic of 18.0 with a p-value of 0.000 and effect size of 0.978 suggests significant differences between the 5,000 demonstration data used instead of the 2,500 data in the demonstration buffer. Similarly, the Mann-Whitney  U-Statistic of 181.0 and a p-value of 0.0000 with an effect size of 0.7737 suggest significant differences between 10,000 and 5,000 demonstration data in the buffer. From the above experiment, we conclude that increasing the number of demonstrations leads to better performance.
              
\subsection{Analysis of Diversity in Demonstration }
From the ablation studies in sub-sections \ref{sec:demo_portion} and \ref{sec:demo_quant}, we conclude that the quantity of demonstrations impacts the performance outcomes than the proportion of demonstrations. This leads us to hypothesize that the diversity in demonstrations is the primary driver of enhanced performance. 

\begin{figure*}[h]
   \centering
   \subfloat[Agent demonstrations\label{agent_demo_density}]{%
      \includegraphics[clip, width=0.5\textwidth]{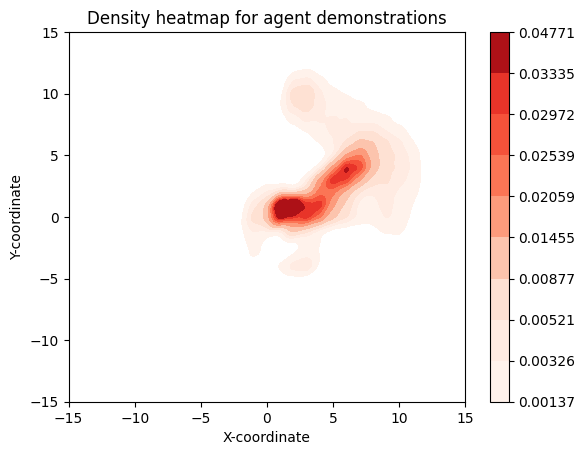}}
   \subfloat[Human demonstrations\label{human_demo_density}]{%
      \includegraphics[clip, width=0.5\textwidth]{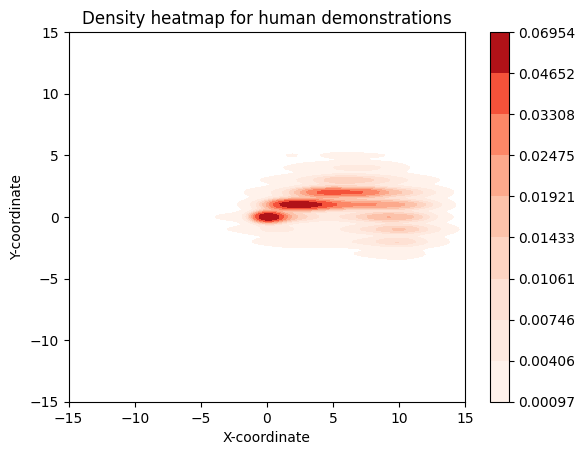}}\\
   \subfloat[Human policy correction\label{policy_correction_density}]{%
      \includegraphics[clip, width=0.5\textwidth]{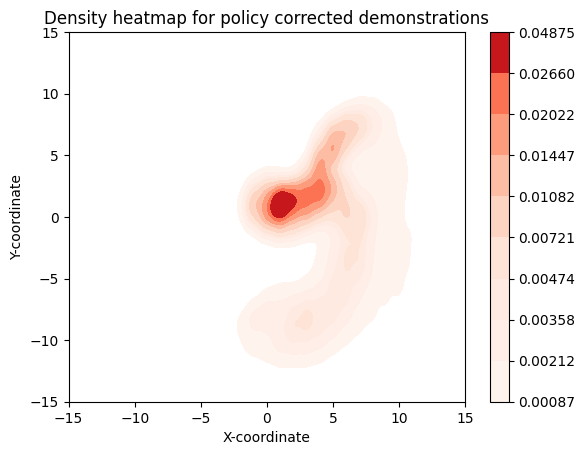}}
   \caption{\label{img:demo_all_density} Density heat map of winning five hundred episodes from trained agents, human users, and human policy corrections. Here, the x-axis and the y-axis represent the drones' scaled relative position.}
\end{figure*}

We investigate the effect of demonstration diversity on the performance of trained agents, guided by agent demonstration, human demonstration, and policy-corrected demonstrations. There's often an abundance of trained agent demonstrations in a simulated real-world complex task, while human demonstrations are scarce due to the high human time and cost. We used 500 winning episodes of human demonstrations from $11$ persons. Similarly, we use 500 winning agents and policy-corrected demonstrations. 

\begin{figure}[h]
  \centering
  \includegraphics[height=0.6\textheight]{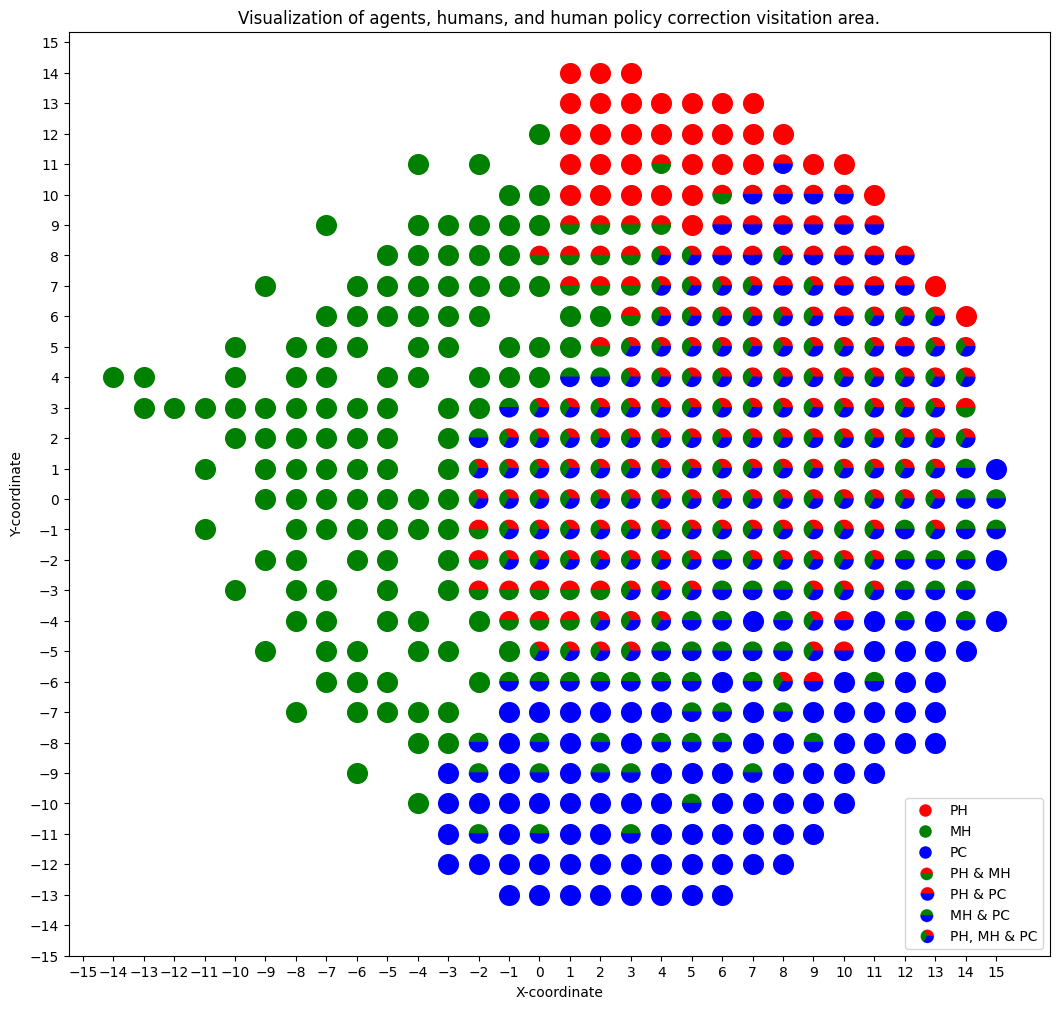} 
  \caption{The area coverage comparison for human demonstrations (HM), agent demonstrations(PH) and human policy correction (PC) with trained agent demonstrations in complex scenario.} 
  \label{img:demo_area_cover}
\end{figure}

\noindent We employ a qualitative approach to assess the diversity of demonstrations by plotting the trajectories of all three demonstration types and examining the spatial coverage within these visualizations. This method allows us to infer the range of strategies and behaviors exhibited across different demonstrations. In our analysis, we generate density heat maps for each type of demonstration, which visually represent the frequency of occurrences at various points in the task space. Specifically, Figure \ref{img:demo_all_density}(a), (b), and (c) correspond to the heat maps for the trained agents, actual human users, and human policy correction demonstrations, respectively. Although our simulated environment is a square space of size $6000 \times 6000$ meter$^2$, we plotted them in reduced space for better visualizations and understanding. The heat maps distinctly reveal that policy-corrected demonstrations show a broader coverage area than those from humans and trained agents. This suggests a greater coverage in the demonstrations collected using policy correction, contributing to the enhanced performance observed earlier in experimental results for this advice-providing technique.

\begin{figure*}[h]
   \subfloat[Area covered by human demonstrations, agent demonstrations and policy-corrected demonstrations]{%
      \includegraphics[clip, width=0.48\textwidth]{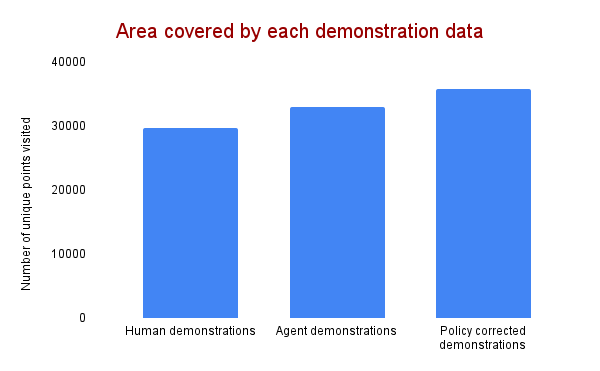}}
    \hspace{\fill}
   \subfloat[Entropy-based diversity measure for human demonstrations, agent demonstrations and policy-corrected demonstrations]{%
      \includegraphics[clip, width=0.48\textwidth]{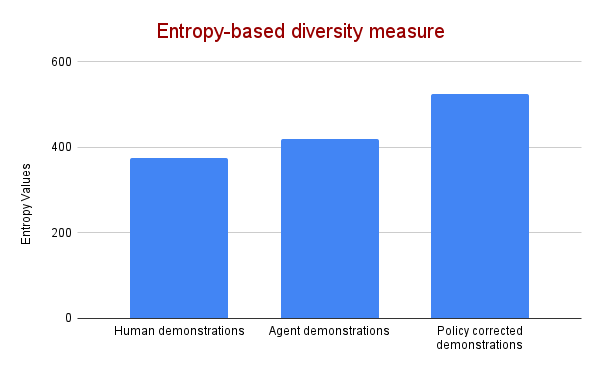}}
\caption{\label{img:diversity_demo_quant}Quantitative measure of diversity in human demonstrations, agent demonstrations and policy-corrected demonstrations. Here, higher entropy values represent more diversity}
\end{figure*}

To better visualize the visitation area (by agents, humans, and human policy correction), we plotted the x and y coordinates of winning demonstrations in a projected space and marked them with different color codes, as shown in Figure \ref{img:demo_area_cover}. The plot clearly shows that the trajectories of humans and trained agent demonstrations overlap less. A diverse range of human actions, from individual to individual, underscores the complexity and variability inherent in human gameplay. In contrast, policy correction effectively uses the strengths of both humans and agents, resulting in a more state-area coverage encompassing all critical zones identified by both. Subsequently, we conducted a quantitative analysis of the area covered by each type of demonstration, the results of which are presented in Figure \ref{img:diversity_demo_quant} (a). From Figure~\ref{img:diversity_demo_quant} (a), we can observe that policy-corrected demonstrations visited 4,116 unique points higher than agent demonstrations and 2,761 points higher than human demonstrations.
\noindent We also employed an entropy-based approach to quantify the diversity of demonstrations, as suggested by Neumann et al. \cite{neumann2020computing}. As shown in Figure \ref{img:diversity_demo_quant} (b), the entropy values for policy-corrected demonstrations were over 10\% higher than those for human and trained agent demonstrations. This empirical evidence confirms our previous observations, indicating that human policy correction introduces significantly diverse states, improving the learning process with various experiences.

\section{Conclusion and Future Work}
In this work, we demonstrated that human-AI collaboration can be better than humans or agents alone in a complex multi-agent task.
By developing a novel simulator and user interface, we have established a platform where humans and AI agents can collaborate and incorporate advice effectively with real-world dynamics. Our experimental result shows that a trained RL agent performs better than a heuristic agent and humans in complex airspace-simulated environments. Our empirical findings underscore the value of incorporating human demonstrations and policy-corrected demonstrations into the training of AI agents, revealing a marked improvement in the agents' learning efficiency and operational performance. In addition, we demonstrated that policy-corrected demonstrations, a human-AI collaboration approach, require less mental demand, temporal demand, and effort, yielding superior performance compared to humans alone. The findings of this study, while significant, are limited by its focus on a single enemy with full and free communication, not fully reflecting the complexity of real-world scenarios.
Future research directions include addressing these gaps by exploring more complex scenarios and diverse human expertise to enhance human-AI collaboration in real-world settings.






\bmhead{Ethics Statement}
Our human subject study was approved by the University's ethics board (REB number: Pro00107555). We have designed the simulator environment so that drones are not equipped with weapons that can directly endanger the lives of humans. We are also focusing on a defensive task to minimize the risk of our work being used by bad actors. 

\bmhead{Declarations} The authors declare that they have no conflict of
interest.

\bmhead{Data availability statement} The datasets generated during and/or analyzed during the current study are available from the corresponding author on request. 












\newpage

\clearpage
\begin{appendices}

\section{Environment design: Team description} \label{app:team_description}

In this section, we discuss the team details of our UVA-based airport security systems shown in Figure~\ref{img:thunderblade}. In the subsequent discussion, we present a mathematical representation of the aforementioned scenario. Let $p_{RA} \in \mathbb{R}^2$ and $r_{RA} \in [0, 1)$ be the center and radius of a circle representing the restricted area, respectively. Two teams are present in this scenario: the ally team (blue team) and the enemy team (red team).

\subsection{Ally team (blue team)}
The blue team comprises five aerial drones, a ground radar (GR) sensor, and a ground control station (GCS). Each ally drone also has several neutralization payloads (i.e., devices capable of neutralizing enemy drones when they are within a certain range). The goal of the blue team is to protect the restricted zone of the airport from the red team by detecting, localizing, and neutralizing the enemy drones. \\

\textbf{Ally Ground Control Station (GCS):} There is one GCS located at position $p_{GC_t} \in \mathbb{R}^2$ at time step $t \in \mathbb{N}$. In addition, let $r_{GC} \in (0, 1)$ denote the circle\text{'s} radius representing the GCS operating range. \\

\textbf{Ally Ground Radar (GR):} Consider $n_{GR} \in \mathbb{N}$ ground radars whose jobs are to gather information about the ally and enemy drones. Denote by $p_{t}^{i,GR} \in \mathbb{R}^2$ the position and by $\phi_{t}^{i,GR} \in [-\pi, \pi]$ the heading angle (orientation) of the $i$-th GR at time $t \in \mathbb{N}$, $i \in \mathbb{N}_{nGR}$. Let $u_{t}^{i,GR} \in [-1, 1]$ denote the action of the $i$-th GR, where $u_{t}^{i,GR}$ is the ratio of angular speed with respect to its maximum value. In particular, $u_{t}^{i,GR}$ controls the rotation of the $i$-th GR as follows:
\begin{equation}
\phi_{t+1}^{i,GR} = \phi_{t}^{i,GR} + v_{GR}^{max} \times u_{t}^{i,GR}
\end{equation}
where $v_{GR}^{max} \in [0, 1)$ is the maximum angular speed. Denote by $\rho_{GR} \in [0, 2\pi]$ the field of view of the GR and by $r_{GR} \in [0, 1)$ the radius of a circle representing its sensing range.

\textbf{Ally Drone (Blue Drone):} The ally system has $n_{AD} \in \mathbb{N}$ drones. The state of the $i$-th ally drone includes:

\begin{enumerate}
    \item Position: $p_{t}^{i} \in \mathbb{R}^2$.
    \item Heading angle: $\phi_{t}^{i} \in [-\pi, \pi]$.
    \item Relative orientation of the electro-optic (EO) sensor: $\phi_{t}^{i,EO} \in [-\phi_{EO}^{\text{max}}, \phi_{EO}^{\text{max}}]$.
    \item Functionality status: $f_{t}^{i} \in \{0, 1\}$, where 0 indicates non-functional.
    \item Control by GCS: $g_{t}^{i} \in \{0, 1\}$, where 0 indicates no control by GCS.
    \item Position of controlling GCS: $p_{t}^{i,GC} \in \mathbb{R}^2$.
    \item Radar status: $\text{radar-enabled}_{t}^{i} \in \{0, 1\}$, where 0 means off.
    \item EMP usage: $\text{emp-used}_{t}^{i} \in \{0, 1\}$, indicating if EMP has been used.
\end{enumerate}

In particular, there is an EMP-auto-destruction probability $pr_{EMPD} \in [0, 1]$ according to which the drone may destroy itself upon using EMP, i.e.,
\begin{equation}
f_{t+1}^{i} = 
\begin{cases} 
0 & \text{if } f_{t}^{i} = 0; \\
1 & \text{if } f_{t}^{i} = 1 \text{ and } emp\text{-}used_t^i = 0; \\
B & \text{if } f_{t}^{i} = 1 \text{ and } emp\text{-}used_t^i = 1;
\end{cases}
\label{eq:17}
\end{equation}
where $B \in \{0, 1\}$ is a Bernoulli random variable with success probability of $1 - pr_{EMPD}$.

Denote by $\rho_{EO} \in [0, 2\pi]$ the field of view of the EO sensor and by $r_{EO} \in [0, 1]$ the radius of a circle representing its sensing range. In addition, let $\rho_{AD} \in [0, 2\pi]$ be the field of view of the radar and $r_{AD} \in [0, 1)$ be the radius of a circle representing its sensing range. The action set of drone $i$ at time $t$ is described below:

\begin{enumerate}
\item Control signal for the movement angle $u_{t}^{i,MA} \in [-\pi, \pi]$,
\item Seed-ratio of the drone, i.e., $u_{t}^{i,SR} \in [0, 1]$ such that
\begin{equation}
p_{t+1}^{i} = p_{t}^{i} + v_{SR}^{max} \times u_{t}^{i,SR} \times \cos(\sin(u_{t}^{i,MA})),
\label{eq:18}
\end{equation}
where $v_{SR}^{max} \in (0, 1)$ is the maximum speed of drone,
\item Angular speed-ratio $u_{i,t}^{heading} \in [-1, 1]$ such that
\begin{equation}
\phi_{t+1}^{i} = \phi_{t}^{i} + v_{AS}^{max} \times u_{t}^{i,heading},
\label{eq:19}
\end{equation}
where $v_{AS}^{max} \in (0, 1)$ is the maximum angular speed,
\item Angular speed-ratio of EO sensor $u_{t}^{i,EO} \in [-1, 1]$ s.t.
\begin{equation}
\phi_{t+1}^{i,EO} = 
\begin{cases} 
-\phi_{EO}^{max} & \text{if } \phi_{t}^{i,EO} \leq -\phi_{EO}^{max}; \\
+\phi_{EO}^{max} & \text{if } \phi_{t}^{i,EO} \geq \phi_{EO}^{max}; \\
\phi_{t}^{i,EO} + v_{ASEO}^{max} \times u_{t}^{i,EO} & \text{otherwise};
\end{cases}
\label{eq:20}
\end{equation}
where $v_{ASEO}^{max} \in (0, 1)$ is the maximum angular speed of EO sensor,
\item Turn off/on the drone’s radar $u_{t}^{i,ER} \in \{0, 1\}$, where 0 refers to turning the radar off, i.e.,
\begin{equation}
\text{radar-enabled}_{t+1}^{i} = u_{t}^{i,ER}.
\label{eq:21}
\end{equation}
\item Turn off/on the EMP $u_{t}^{i,EMP} \in \{0, 1\}$, where 0 refers to turning the EMP off, i.e.,
\begin{equation}
\text{emp-used}_{t+1}^{i} = u_{t}^{i,EMP} + \text{emp-used}_{t}^{i}(1 - u_{t}^{i,EMP}).
\label{eq:22}
\end{equation}
\item Turn off/on jamming, i.e., $u_{t}^{i,EJ} \in \{0, 1\}$, where 0 refers to turning the jamming off.
\item Turn off/on GPS spoofing, i.e., $u_{t}^{i,GPSS} \in \{0, 1\}$, where 0 refers to turning the spoofing off.
\item Turn off/on hacking, i.e., $u_{t}^{i,EH} \in \{0, 1\}$, where 0 refers to turning the hacking off.
\end{enumerate}

\subsection{Enemy team (Red team)}
The red team comprises a single drone equipped with its own radar sensor and a potentially hazardous payload. The enemy team consists of $n_{EGCS} \in \mathbb{N}$ GCSs and $n_{ED} \in \mathbb{N}$ drones.

\textbf{Enemy Ground Control Station (GCS):} The position of the $i$-th enemy GCS at time $t$ is $p_{i}^{EGCS}(t) \in \mathbb{R}^2$. The radius of its operating range is denoted by $r_{EGCS} \in [0, 1)$. \\

\textbf{Enemy Drone (Red Drone):} The state of the $i$-th enemy drone at time $t$ includes:

\begin{enumerate}
    \item Position: $p_{t}^{i,ED} \in \mathbb{R}^2$.
    \item Payload: $l_{t}^{i,ED} \in \{0, 1, 2, 3\}$, where values represent "safe", "unknown", "moderate", and "dangerous", respectively.
    \item Control by GCS: $g_{t}^{i,ED} \in \{0, 1\}$.
    \item Position of controlling GCS: $p_{t}^{i,EGCS} \in \mathbb{R}^2$ where 0 means the drone is not controlled by the GCS
    \item Functionality status: $f_{t}^{i,ED} \in \{0, 1\}$ where 0 means it is not functional; In particular,
    \begin{equation}
    f_{t+1}^{i,ED} = 
    \begin{cases} 
    0 & \exists j \in N_{nAD} \text{ s.t. } f_{t}^{j} \times u_{t}^{j} = 1 \text{ and } \| p_{t}^{i,ED} - p_j \|_2 \leq r_N, \\
    0 & \| p_{t}^{i,ED} - p_{t}^{i,EGCS} \|_2 > r_{EGCS} \text{ and } gt_{t}^{i,ED} = 1, \\
    ft_{t}^{i,ED} & \text{otherwise}.
    \end{cases}
    \label{eq:23}
    \end{equation}
    where $r_N \in [0, 1)$ is the radius of a circle representing the neutralization range of ally drones and $u_j^t \in \{0, 1\}$ is defined as whether or not the ally drone $j \in N_{nAD}$ is enabled to neutralize enemy drone $i$, i.e.,
    \begin{equation}
    u_{t}^{j} := u_{t}^{j,EMP} \lor u_{t}^{j,GPSS} \lor (gt_{t}^{i,ED} \land u_{t}^{j,EJ}) \lor (gt_{t}^{i,ED} \land u_{t}^{j,EH}).
    \label{eq:24}
    \end{equation}
\end{enumerate}

\section{Experimental Platform and Architecture for HitL Interactions}\label{app:hilt_user-interface}
\begin{figure}[h]
  \centering
  \includegraphics[width=0.9\columnwidth]{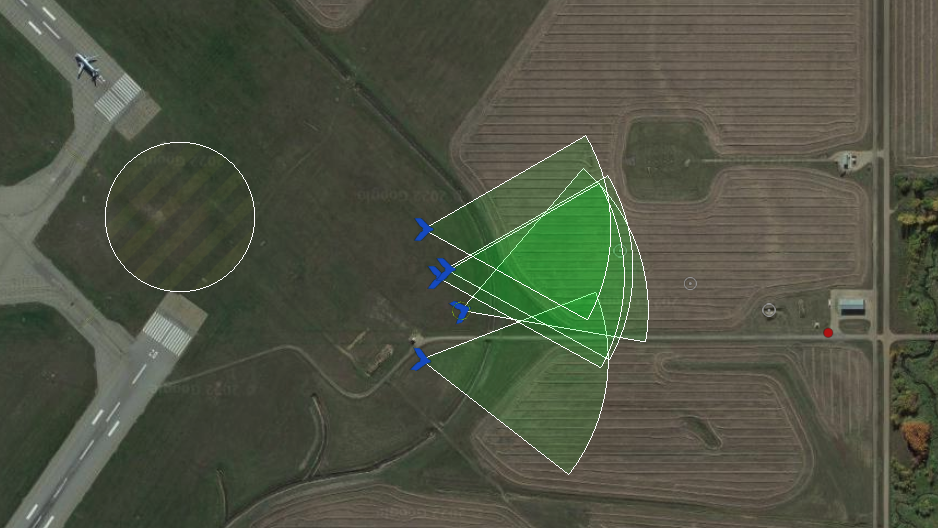}
  \caption{User interface for human operators to control the agents}
  \label{img:user_interface}
\end{figure}

The experimental platform is a distributed application built using the Cogment platform. We use Cogment as it simplifies the conduction of large-scale experiments in multi-agent systems and HitL. Cogment dispatches observations of the environment from the simulation to the agents, as well as instructions from higher-level agents such as humans or decision-makers. It then dispatches agents' actions to
the environment, which updates the simulation and agent's
instructions. Furthermore, a priority-ordered list comprising multiple agents can be allocated simultaneously to a single drone entity. When an agent with higher priority issues a command for velocity or rotation change, it overrides those from lower-priority agents. This mechanism facilitates dynamic takeover by the human operator, optimizing communication channels, data storage, and processing pipelines in the process.

\begin{figure}[h]
  \centering
  \includegraphics[width=0.95\columnwidth]{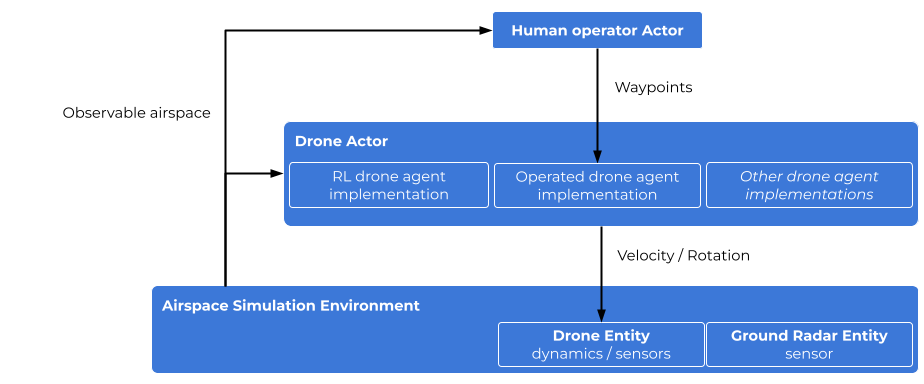}
  \caption{Airspace simulation and hierarchical multi-agent modeling} 
  \label{img:thunderbladeenv}
\end{figure}

In order for the human operator to control the ally drones and online policy correction, we have developed a user interface as a part of the experimental platform shown in Figure~\ref{img:user_interface}. The experimental platform is built around a simplified airspace simulator operating in 2D, simulating two types of entities: drones from both blue and red teams and the ground radar on the blue side. While simplified, several aspects have been modeled following real-world specifications provided by defense experts, such as the detection capabilities of the drone sensors and the radar, as well as the dynamics of the fixed-wing drones. The communication between the different agents is limited due to their partial observation of the environment but is perfect and instant. The experimental platform models the scenarios as a multi-agent system with a three-layer architecture. The primary agent type controls the drones through velocity and rotation changes. 
The bottom layer is named the drone agent layer, and it receives partial observation of the environment "through the lens" of its sensors. In this layer, the blue drones can be either fully autonomous, fully human-operated, or hybrid (i.e., control is shared by the human-agent team). This is implemented by two Cogment actor implementations (drone and human actors), shown in Figure-\ref{img:thunderbladeenv}. For each ally drone agent in the simulation, Cogment instantiates two actors using those implementations; it can then dynamically assign the control of the drone entity to one of them.

The system supports two types of high-level decision-makers: the Control and Command agents. The Control Agent simplifies drone-level tasks by assigning targets, a capability that can be leveraged by either a human or an agent. Meanwhile, an additional layer is incorporated to allow the Command Agent to designate areas of interest to the Control Agents, facilitating more nuanced and strategic operations.
Human operators were introduced to facilitate the control of individual drones and to gather demonstrations of various drone behaviors. These operators can select specific drones and set waypoints for them by predicting the anticipated trajectories of enemy drones. In Figure~\ref{img:user_interface}, the waypoints are denoted by grey circles on the map.  Once a specific waypoint has been defined, Cogment dynamically gives control of the associated ally drone to the operated agent, causing it to move towards the defined waypoint via the shortest path under the standard physical dynamics constraints. Similarly, the human operator can delete existing waypoints through the interface to rectify any errors in predicting the enemy drone trajectory. Conversely, when no waypoints are defined, the control is given back to the autonomous agent. The interface also allows the human to operate at $3$ simulation speeds {1x, 2x, 5x}, according to their preferences.

Figure \ref{img:thunderbladeenv} represents the architecture of the developed experimental platform. The Cogment platform \cite{cogment} handle the orchestration of the execution and communication between the different components:

\begin{enumerate}
   \item The drone agents can each use one of multiple implementations; they are encapsulated in dedicated micro services as Cogment actors.
   \item The simulation, encapsulated in a dedicated micro service as a Cogment environment (that uses MDP formalism).
   \item The human operator, interacting through the UI, encapsulated as a client Cogment actor. 
   \end{enumerate}

The experimental platform along with a user interface is developed for this project enabled the team to easily implement a heterogeneous, hierarchical multi-agent system. This allowed the integration of multiple types of agents, each with their own specialized capabilities and roles, within a single system while the hierarchical properties enabled task decomposition. Furthermore, a priority ordered list of multiple agents can be assigned to a single drone entity at once. 
One key feature of the platform is the dynamic agent or human \textquotedblleft takeover\textquotedblright capability, which supports human-AI teaming during operations and provides advice during training. This allows for the seamless integration of human operators and AI agents within the system, allowing the operator to take over when needed. This allowed us to evaluate the human-AI team. A detailed description of the user interface can be found in Section \ref{app:user_interface}.

\section{Hyperparameters}\label{app:hyper-params}
We perform hyper-parameter tuning using grid search to find best parameters for the algorithms and use the same settings for all the models in scenarios. We considered learning rate values of[$0.4$, $0.04$, $0.004$, $0.0004$, $0.00004$], epsilon decay values of [$0.99$, $0.995$, $0.9995$, $0.99995$, $0.999995$], and discount factor values of [$0.9$, $0.99$, $0.999$]. The supervised loss coefficient weight ($\lambda_2$ in equation~\ref{eqn:ddqn_demo}) varied between  $10^{5}$ to $1$, and we set $\lambda_3$ to 0. The same network structure was used across D3QN-0-2500, D3QN-0-500, D3QN-500-2000, D3QN$_{HM}$-500-0 and D3QN$_{PC}$-500-0, consisting of two hidden layers with $64$ fully connected neurons. A final fully connected layer was added to represent each action's Q-values. The non-linearity function used in all layers was rectified linear units (ReLU). During training, we use the Adam optimizer and applied an epsilon-greedy policy, gradually reducing epsilon from $1$ to $0.05$. The batch size was $64$, and the replay memory size was $100,000$.

\begin{table}[!h]
\centering
    \begin{tabular}{ |c|c| } 
    \hline
    Parameter & Value \\ [0.5ex] 
    \hline\hline
        Training Episodes & 10,000 \\ 
        Replay Memory size & 100,000 \\ 
        Batch size & 64 \\
        Learning rate & 0.0004\\
        Discount factor & 0.99\\
        Target network update frequency & 10 \\ 
        Initial $\epsilon$ & 1.0 \\
        Final $\epsilon$ & 0.05 \\
        $\epsilon$ decay per episode &  0.999995\\
    \hline
    \end{tabular}
\caption{\label{hyperparameter-table} Model hyper-parameters.}
\end{table} 

\section{Demonstration visualization} \label{app:demo_visualization}

\begin{figure*}[h]
    \centering
   \subfloat[trained agent demonstration\label{human_demo}]{%
      \includegraphics[clip, width=0.5\textwidth]{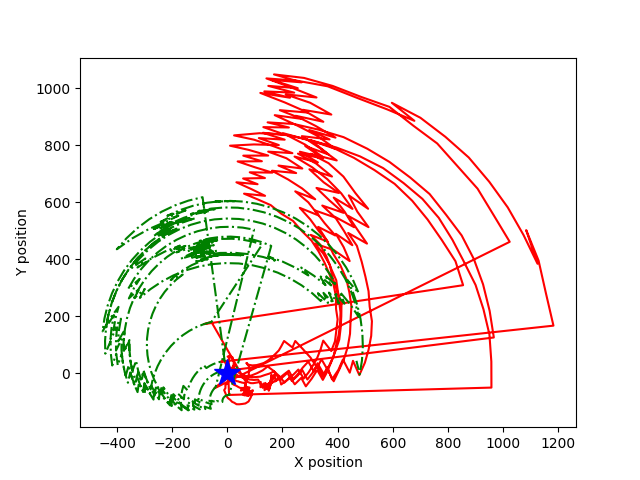}}
   \subfloat[a human user (User~$1$)\label{pyramidprocess} ]{%
      \includegraphics[clip, width=0.5\textwidth]{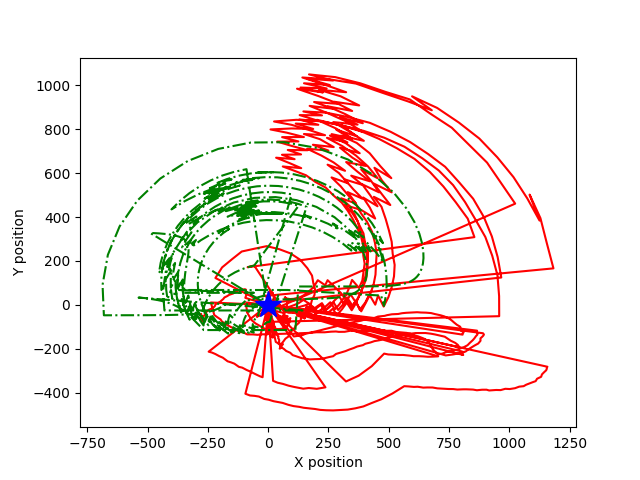}} \\
   \subfloat[a human user (User~$2$)\label{mt-simtask}]{%
      \includegraphics[clip, width=0.5\textwidth]{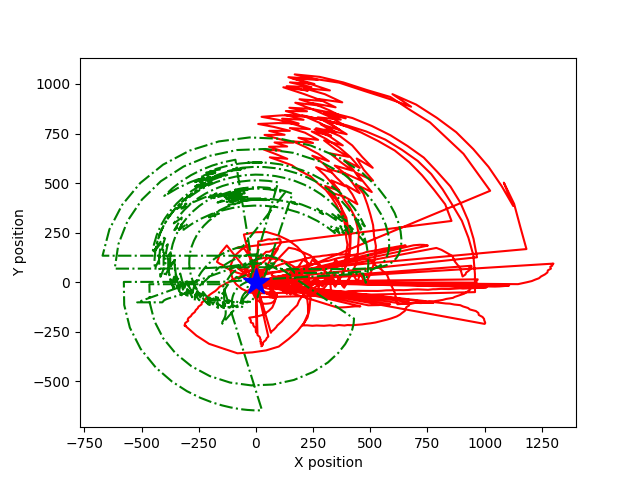}}
\caption{\label{img: human_demo}Visual representation of five episodes from trained agent and two different real human users}
\end{figure*}

We visualized five trajectories of all blue and red drones from trained agent demonstrations and two actual human demonstrations from different users who played more than $30$ games, as shown in Figure~\ref{img: human_demo}. In this figure, the blue star denotes one of the ally drones that neutralized the enemy drone and is the frame of reference (located at $(0,0)$). The red and green lines represent the relative position of the enemy drone and the restricted airspace (with respect to the blue drone's position). Figure~\ref{img: human_demo}(a) shows five trajectories of ally drones generated from the trained D3QN agent. 
We note that across all the figures, the red drone starts moving towards the restricted zone
while being chased by the blue drones until it is neutralized. The low density of red lines around the blue star indicates that the blue drones quickly neutralize the red drone without following it for a long time.  
From Figure~\ref{img: human_demo}(b) and ~\ref{img: human_demo}(c), which depicts trials by two different human participants, we notice that there is more movement (high-density) around the blue star, suggesting that the human tries setting waypoints in different areas (using the whole team of five blue drones) of the map to neutralize the red drone. These trajectories are sub-optimal (longer trajectory length) as compared to the trajectories from trained agent demonstrations. However, these might be helpful to neutralize the red drones in challenging environment configurations where the trained RL agents fail to catch the enemy drone (trained agents have a failure rate of around $10\%$ in this task).

\section{NASA-TLX Load Index Questionnaires:} \label{app:nasa_tlx}
Participants will complete a 6-question NASA-TLX workload assessment, with each question featuring a 21-point slider ranging from \textquotedblleft very low\textquotedblright{}  to \textquotedblleft very high\textquotedblright{} . These questions are modified to facilitate comparison with the previous round of the task:
\begin{enumerate}[label=\arabic*.]
    \item Mental Demand: How mentally demanding was the task compared to the previous round?
    \item Physical Demand: How physically demanding was the task compared to the previous round?
    \item Temporal Demand: How hurried or rushed was the pace of the task compared to the previous round?
    \item Performance: How successful were you in accomplishing what you were asked to do compared to the previous round?
    \item Effort: How hard did you have to work to accomplish your level of performance compared to the previous round?
\end{enumerate}

\textbf{Demographic Questionnaires:}
\begin{itemize}
    \item Gender? [Multiple Choice: Woman, Man, Transgender, Prefer to describe myself, Prefer not to respond]
    \item Your current position/post
    \item What is your age? [slider 18-65]
    \item Are you experienced in defence strategies/drone control, development? [Yes/No]
    \item Years of experience in  defence strategies/drone control, development? [slider or box]
    \item Did you experience drone controls in real life or simulated tasks in the past ? [yes/no]
    \item Do you have experience with video games? [Yes/No]
        If yes, years of experience with video games.
\end{itemize}




\end{appendices}


\newpage
\bibliography{sn-bibliography}

\end{document}